\documentclass[10pt,twocolumn,letterpaper]{article}

\usepackage[pagenumbers]{cvpr} 

\usepackage{algorithm}
\usepackage[noend]{algorithmic}
\usepackage{makecell}
\usepackage{booktabs} 
\usepackage{multirow} 
\usepackage[table]{xcolor}
\usepackage{bbding}

\definecolor{cvprblue}{rgb}{0.21,0.49,0.74}
\usepackage[pagebackref,breaklinks,colorlinks,allcolors=cvprblue]{hyperref}
\usepackage{placeins}
\def\paperID{} 
\def\confName{CVPR}
\def\confYear{2026}

\title{Unified Spatiotemporal Token Compression for Video-LLMs \\ at Ultra-Low Retention}

\author{Junhao Du$^{1}$\thanks{Equal contribution.} \quad Jialong Xue$^{1}$\footnotemark[1] \quad Anqi Li$^{1}$ \quad Jincheng Dai$^{2}$ \quad Guo Lu$^{1}$\thanks{Corresponding author.} \\ $^{1}$Shanghai Jiao Tong University \quad $^{2}$Beijing University of Posts and Telecommunications\\ {\tt\small \{junhao\_du, mengdeerer, anqi.li, luguo2014\}@sjtu.edu.cn \quad daijincheng@bupt.edu.cn} }

\begin{document}
\maketitle

\begin{abstract}
Video large language models (Video-LLMs) face high computational costs due to large volumes of visual tokens. Existing token compression methods typically adopt a two-stage spatiotemporal compression strategy, relying on stage-specific metrics and an implicit assumption of spatiotemporal separability. Under extremely low retention ratios, however, such approaches often result in unbalanced allocation and loss of visual evidence essential for question answering. We reformulate token compression as a spatiotemporal allocation task within a global token retention pool. We propose a unified selection mechanism that integrates attention weights and semantic similarity to globally select tokens with high contribution and low redundancy. Unselected tokens are merged via clustering and refilled, preserving information integrity. Inside the LLM, we further introduce text-aware merging to perform secondary compression based on query relevance. Without requiring retraining, our method serves as a plug-and-play module compatible with existing Video-LLMs. Experiments show that retaining only about 2\% of visual tokens preserves 90.1\% of baseline performance across multiple benchmarks, while reducing FLOPs to roughly 2.6\%. These benefits generalize across diverse backbones, decreasing end-to-end inference latency and memory consumption. Our unified spatiotemporal token compression strategy establishes the state-of-the-art in video understanding under ultra-low token retention.
\end{abstract}

\begin{figure}[t]
    \centering

    \begin{subfigure}[b]{0.49\linewidth} 
        \centering
        \includegraphics[width=\linewidth]{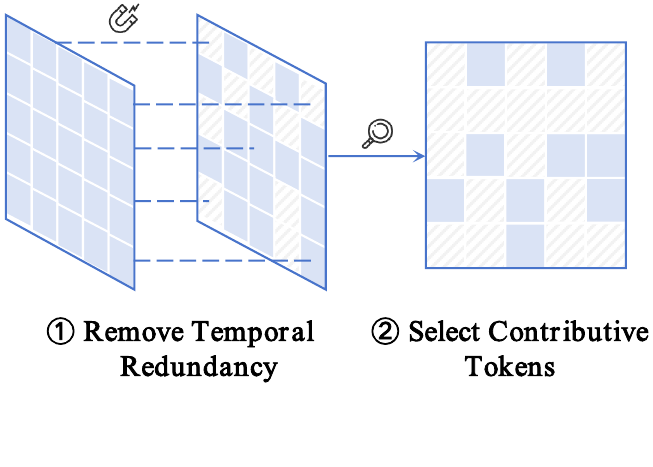}
        \captionsetup{font=scriptsize}
        \caption{Two-Stage Token Compression}
        \label{fig:new-left}
    \end{subfigure}
    \hfill 
    \begin{subfigure}[b]{0.49\linewidth} 
        \centering
        \includegraphics[width=\linewidth]{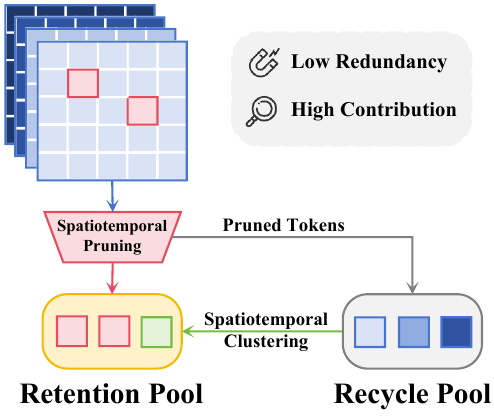}
        \captionsetup{font=scriptsize}
        \caption{Unified Spatiotemporal Compression}
        \label{fig:new-right}
    \end{subfigure}

    \vspace{5pt}

    \begin{subfigure}[b]{1\linewidth}
        \centering
        \includegraphics[width=\linewidth]{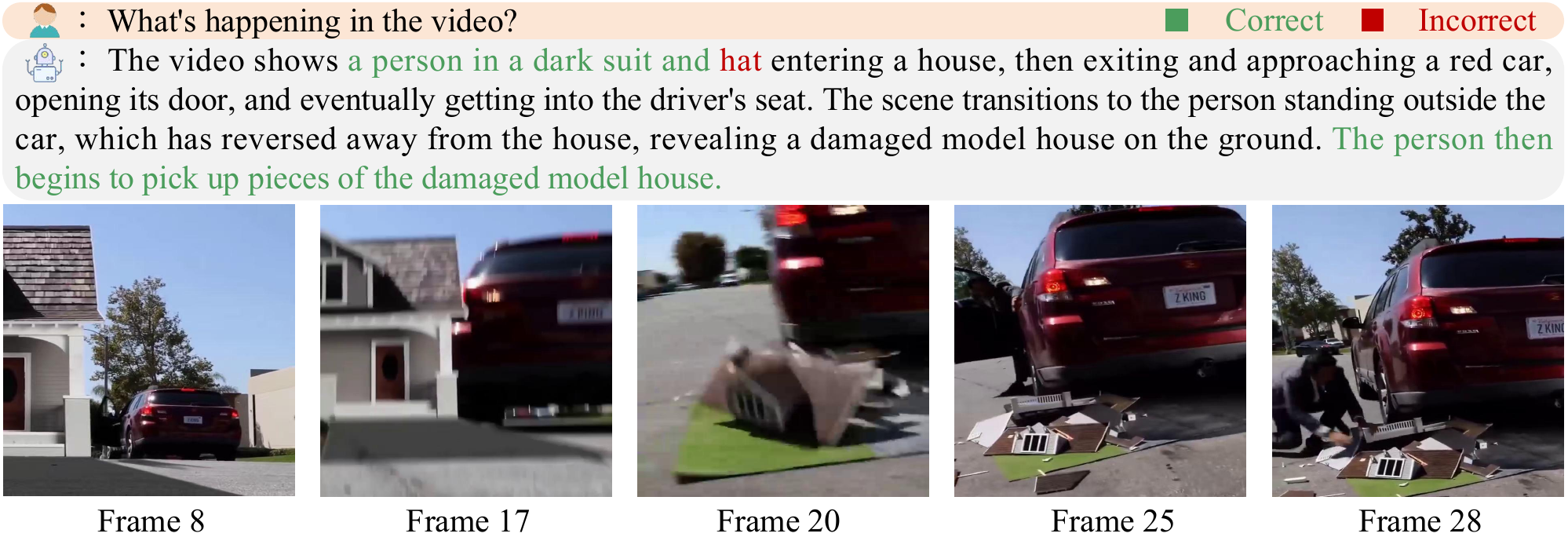}
        \caption{LLaVA-OneVision-7B for Question-Answering}
        \label{fig:introshow-original}
    \end{subfigure}

    \vspace{5pt}
    
    \begin{subfigure}[b]{0.48\linewidth}
        \centering
        \includegraphics[width=\linewidth]{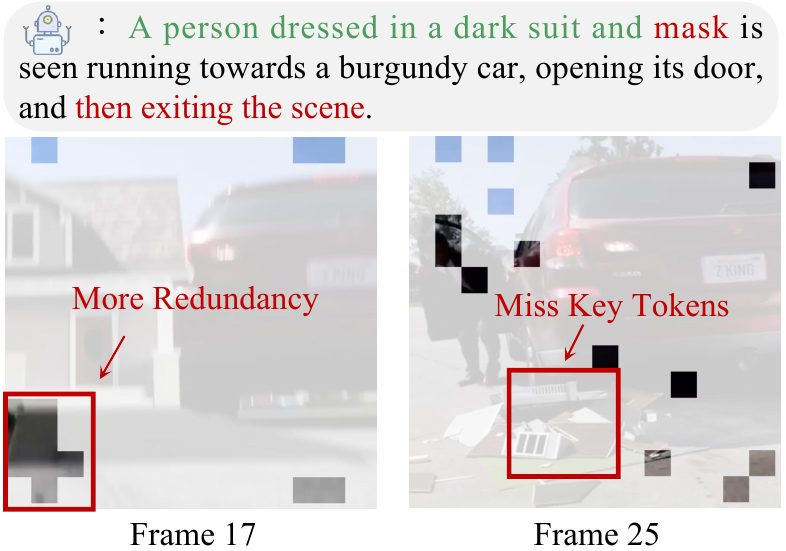}
        \caption{Selected Tokens by HoliTom}
        \label{fig:introshow-holitom}
    \end{subfigure}
    \hfill
    \begin{subfigure}[b]{0.48\linewidth}
        \centering
        \includegraphics[width=\linewidth]{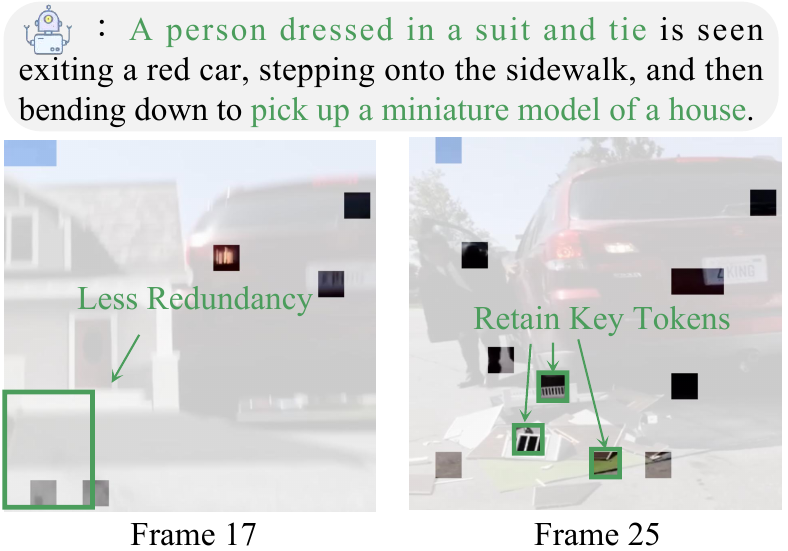}
        \caption{Selected Tokens by Ours}
        \label{fig:introshow-ours}
    \end{subfigure}
    
    \caption{\textbf{(a) Previous method} remove redundant tokens via separate spatial and temporal stages. \textbf{(b) Our method} maintains a global retention pool for unified spatiotemporal redundancy removal. \textbf{(c) Response of LLaVA-OneVision-7B} to video content understanding. \textbf{(d) At 5\% token retention, HoliTom} retains redundant tokens and misses critical ones, leading to misunderstanding (\textcolor{red}{red}). \textbf{(e) Our method} effectively mitigates these issues and yields the correct response (\textcolor{green!60!blue}{green}).}
    \label{fig:introshow}
    \vspace{-10pt}
\end{figure}

\section{Introduction}
\label{sec:intro}

Video large language models (Video-LLMs)~\cite{chen2024expanding,cheng2024videollama,li2024llava,li2023videochat,li2024llama,wang2024qwen2,xu2024pllava,zhang2023video,zhang2024video,bai2025qwen2,wang2025internvl3,yu2025minicpm} have exhibited remarkable performance in complex video understanding tasks. However, their practical deployment remains challenging due to high computational costs and substantial resource demands. This bottleneck is driven primarily by the large volumes of visual tokens. For instance, in the representative model LLaVA-OneVision-7B~\cite{li2024llava}, a single video frame generates 196 visual tokens. A 32-frame video segment thus accumulates up to 6,272 tokens, many of which are highly redundant and information sparse. Even when token retention is reduced to 10\%, hundreds of tokens persist, resulting in persistently high inference latency and memory consumption.

Current training-free video token compression methods primarily follow three technical pathways. 1) Spatial dimension pruning reduces tokens by retaining visually salient or high-attention regions, as demonstrated by attention-based methods such as VisionZip~\cite{yang2025visionzip} and PruMerge~\cite{shang2024llava}. 2) Temporal pruning and merging techniques, exemplified by DyCoke~\cite{tao2025dycoke} and TempMe~\cite{shen2024tempme}, eliminate temporal redundancy by identifying and merging semantically similar tokens across frames. 3) Staged spatiotemporal methods like FastVid~\cite{shen2025fastvid} and HoliTom~\cite{shao2025holitom} implement sequential filtering along both temporal and spatial dimensions. These approaches generally rely on independent scoring strategies for each stage or dimension, implicitly assuming that spatiotemporal redundancy in video tokens is separable. However, under extremely low token retention rates ($\leq 5\%$), this assumption frequently breaks down. As illustrated in~\cref{fig:introshow}, staged decision-making mechanisms tend to allocate spatiotemporal resources unevenly, resulting in the preservation of non-essential tokens while discarding critical ones. For instance, FastVid~\cite{shen2025fastvid} achieves only 83.3\% of its original performance at a 2\% retention ratio. Moreover, within LLMs, most existing methods, such as FastV~\cite{chen2024image}, PDrop~\cite{xing2024pyramiddrop}, and HoliTom~\cite{shao2025holitom}, utilize only the last token's attention weights as the selection criterion. This practice not only introduces positional bias but also diminishes the semantic influence of key query terms, ultimately constraining the retention of meaningful information during the compression process.

We reformulate video token compression as a spatiotemporal token allocation problem under a global constraint, aiming to select tokens that maximize informational contribution while minimizing semantic redundancy. Unlike prior staged approaches, the proposed unified spatiotemporal compression mechanism enables a globally balanced trade-off between token contribution and redundancy. To this end, we design a selection module that integrates visual attention scores with semantic similarity measures, performing joint spatiotemporal evaluation across all tokens. This process identifies key tokens with high contribution and low redundancy for inclusion in the retention pool. Tokens not selected are transferred to a recycle pool, where they undergo semantic-aware aggregation via clustering to preserve the overall semantic structure. Furthermore, during the LLM’s internal processing stage, a query-guided text-aware merging strategy is introduced to perform secondary compression on token representations, effectively retaining query-relevant context. The entire module operates without training and ensures plug-and-play compatibility, allowing flexible integration into existing Video-LLMs.

Our proposed approach is integrated into several mainstream Video-LLM architectures and evaluated on standard benchmarks. Results show that even under extreme compression settings where only 2\% video tokens are retained, corresponding to an average of about four tokens per frame, the model preserves approximately 90\% of its original performance. At the same time, the significant reduction in redundant tokens greatly decreases computational overhead, with FLOPs reduced to about 2.6\% of the original amount. 
Furthermore, our approach achieves superior performance across a diverse range of Video-LLM architectures, demonstrating strong generalizability and deployment convenience.

Our contributions are summarized as follows:

\begin{itemize}
    \item We achieve video token compression using a unified spatiotemporal pool to maximize informational contribution and minimize semantic redundancy.
    \item Even under ultra-low token retention ratios, our method preserves most of the original performance while significantly boosting efficiency, reaching SOTA competency.
    \item We design the entire module to be training-free and plug-and-play, ensuring flexible integration into existing Video-LLMs.
\end{itemize}

\section{Related Work}
\label{sec:relatedwork}

\noindent\textbf{Video Large Language Models.}
The deep integration of pretrained LLMs with video encoders has facilitated the emergence of Video-LLMs, which possess robust capabilities in video understanding and question answering. For instance, VideoChat~\cite{li2023videochat} and VideoLLaMA~\cite{cheng2024videollama} extract video representations through video encoders to image-based LLMs and are further trained on video datasets to enhance their video comprehension abilities. Meanwhile, LLaVA-NeXT~\cite{li2024llavanext}, LLaVA-OneVision~\cite{li2024llava}, and Qwen2.5-VL~\cite{bai2025qwen2} demonstrate exceptional multimodal performance in high-resolution image, multi-image, and video scenarios. However, the extensive number of visual tokens corresponding to long video inputs severely constrains the inference efficiency of Video-LLMs. Although models such as LLaVA-OneVision~\cite{li2024llava}, LLaVA-Video~\cite{zhang2024video}, and InternVL~\cite{wang2025internvl3} attempt to reduce the number of visual tokens through pooling or pixel unshuffle operations, the tokens transmitted to downstream language models still number in the thousands, resulting in substantial computational overhead. While methods like VILA~\cite{lin2024vila}, NVILA~\cite{liu2025nvila}, and MiniCPM-V~4.5~\cite{yu2025minicpm} optimize token usage efficiency, they rely on large-scale model finetuning or training, which incurs high adaptation costs. 
Therefore, the development of an efficient and training-free visual token compression method is critical for enabling efficient inference in Video-LLMs without introducing additional training overhead.
\begin{figure*}[t]
    \centering
    \includegraphics[width=1\linewidth]{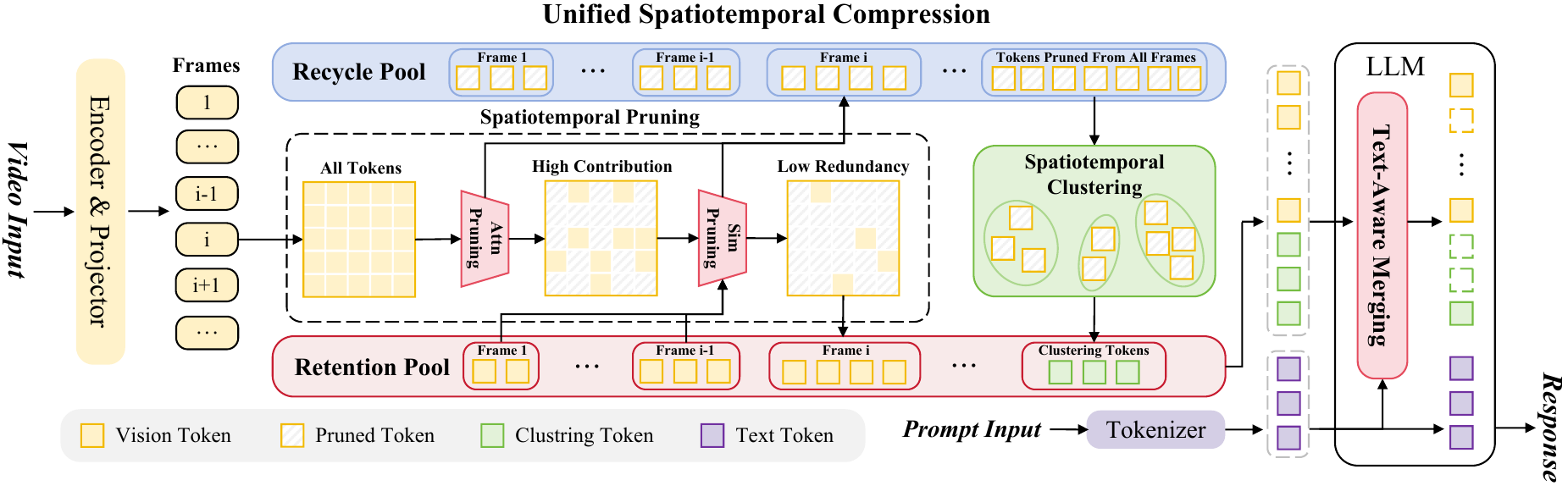}
    \caption{\textbf{Overview of our method.} \textbf{1) The Unified Spatiotemporal Compression} module filters tokens with high contribution and low semantic redundancy, incorporating them into the retention pool. At the same time, it performs clustering and merging on tokens in the recycle pool to preserve the integrity of visual semantic information. \textbf{2) The Text-Aware Merging} mechanism further enhances answer accuracy by guiding the LLM to focus on visual tokens that are most relevant to the input query.}
    \label{fig:overview}
    \vspace{-5pt}
\end{figure*}

\noindent\textbf{Visual Token Compression.}
Token compression has been widely adopted as a strategy to reduce redundancy in both ViTs and LLMs. Existing approaches can be broadly categorized based on their handling of spatial, temporal, and joint spatiotemporal dimensions. Spatial compression operates within individual frames. ToMe~\cite{bolya2022tome} merges similar tokens in ViTs, while LLaVA-PruMerge~\cite{shang2024llava} and VisionZip~\cite{yang2025visionzip} utilize attention weights to eliminate spatial redundancy.
Temporal compression targets redundancy across frames. TempMe~\cite{shen2024tempme} merges redundant temporal tokens in adjacent frames, DyCoke~\cite{tao2025dycoke} dynamically fuses similar tokens, and PruneVID~\cite{huang2024prunevid} clusters redundant ones.
Joint spatiotemporal compression processes the two dimensions in separate stages. Methods such as TESTA~\cite{ren2023testa} perform token fusion across both temporal and spatial dimensions, achieving up to 75\% token reduction. FastVID~\cite{shen2025fastvid} integrates dynamic partitioning with spatiotemporal merging for efficient compression, and LLaVA-Scissor~\cite{sun2025llavascissor} constructs a connectivity graph to select a set of non-overlapping tokens that preserve essential semantic information. HoliTom~\cite{shao2025holitom} utilizes dynamic programming for frame segmentation and proposes a globally redundancy-aware merging mechanism.
Other strategies, such as DivPrune~\cite{alvar2025divprune}, DART~\cite{wen2025dart}, and VisPruner~\cite{zhang2025vispruner}, enhance diversity and preserve salient tokens under high pruning ratios.
In terms of pruning within LLMs, FastV~\cite{chen2024image} performs training-free compression by removing non-essential visual tokens in early layers based on attention scores. 
PDrop~\cite{xing2024pyramiddrop} adopts a progressive pruning schedule for gradual compression. 
TopV~\cite{yang2025topv} formulates token pruning as an optimization problem to identify and eliminate redundancy, while DyMU~\cite{wang2025dymu} dynamically adjusts the pruning ratio according to image complexity. SparseVLM~\cite{zhang2024sparsevlm} introduces a token recycling mechanism.
Although these methods achieve competitive performance under moderate compression ratios, they still exhibit limitations in fully eliminating redundant video tokens. Considerable potential for improvement remains, particularly in scenarios that require ultra-low token retention for Video-LLMs.

\section{Method}
\label{sec:method}

\subsection{Overview}
As depicted in \cref{fig:overview}, our proposed framework comprises two core components: the unified spatiotemporal token compression strategy applied externally to the LLM, and the text-aware token merging mechanism operating internally within the LLM. 
The external compression phase identifies and retains tokens with low redundancy, high contribution, and semantic completeness. This directly reduces the token count to meet the target retention ratio.
Internally, the text-aware merging mechanism prioritizes visually grounded tokens with strong textual relevance based on attention weights and semantic correlations, while consolidating less relevant tokens into their semantically nearest preserved counterparts. By synergistically integrating these two strategies, the method achieves efficient token compression even under ultra-low visual token retention ratios.

\subsection{Unified Spatiotemporal Compression}

Referring to~\cref{fig:overview}, visual token compression uses a retention pool and a recycle pool. Initially, high-contribution, low-redundancy tokens are selected into the retention pool based on attention scores and cosine similarity. The remaining tokens are put in the recycle pool, where they undergo merging via k-nearest neighbors (DPC-KNN)~\cite{du2016study, rodriguez2014clustering}. Finally, these merged tokens are reintroduced into the retention pool, achieving significant compression while maintaining semantic structure.

\noindent\textbf{Spatiotemporal Pruning.}~Drawing on existing approaches~\cite{shang2024llava, wang2024cls, zhang2024cls, yang2025visionzip}, we utilize the attention scores of the CLS token to quantify the contribution of each visual token to the LLM output. The specific computation is defined as follows:
\begin{equation}
    \boldsymbol{A}_h=\mathrm{Softmax}\left(\frac{\boldsymbol{Q}_h\boldsymbol{K}_h^\top}{\sqrt{d}}\right),
    \label{eq:attn}
\end{equation}
where $\boldsymbol{A}_h$ is the attention score of each head, $\boldsymbol{Q}_h$ and $\boldsymbol{K}_h$ represent query and key, respectively. Taking the average over the head dimension yields the average attention matrix $\boldsymbol{A}_{avg}$, reflecting how each token attends to other tokens.
For visual encoders such as SigLIP~\cite{zhai2023siglip} that lack an explicit CLS token, we compute the average attention score between each visual token and all other tokens in the sequence as an equivalent CLS attention score. Based on these scores, the top $k$ tokens with the highest attention values are selected. 

We then calculate the cosine similarity between each candidate token and those already present in the retention pool to evaluate semantic redundancy, using the following formula:
\begin{equation}
    \boldsymbol{S}=\mathrm{sim}(c,\mathcal{P})=\max_{p\in \mathcal{P}}\frac{c\cdot p}{\|c\|\|p\|},
\end{equation}
where $c$ denotes the candidate token and $\mathcal{P}$ is the set of tokens currently in the retention pool. 
A candidate token $c$ is incorporated into the retention pool only if its maximum similarity to any token in $\mathcal{P}$ remains below a predefined threshold $\tau$; otherwise, it is put in the recycle pool. This iterative selection process continues until the number of tokens in the retention pool reaches the predetermined capacity limit.

\noindent\textbf{Spatiotemporal Clustering.}~To maintain the semantic integrity of the visual tokens preserved in the retention pool, tokens assigned to the recycle pool are merged using DPC-KNN. Specifically, given a set of tokens $\{v_1, v_2, ..., v_N\}$ in the recycle pool, the algorithm evaluates the potential of each token $v_i$ to act as a cluster center by computing its local density $\rho_i$ and its minimum distance $\delta_i$ to any token with higher density. These two quantities are defined as follows:
\begin{equation}
    \rho_i=\exp\left(-\frac{1}{k}\sum_{v_j\in\mathrm{kNN}(v_i)}d(v_i,v_j)^2\right),
\end{equation}
\begin{equation}
    \delta_i=
\begin{cases}
\max_{j\neq i}  d(v_i,v_j) & \mathrm{if~}\rho_i=\max_k\rho_k \\
\min_{j:\rho_j>\rho_i}  d(v_i,v_j) & \mathrm{otherwise}
\end{cases},
\end{equation}
where $d(\cdot,\cdot)$ denotes the Euclidean distance.
Based on the decision scores $\gamma_i = \rho_i \times \delta_i$, tokens with the highest $\gamma$ values are selected as cluster centers. The remaining tokens are then assigned to their nearest cluster center according to feature-space distances. The features of all tokens belonging to the same cluster are averaged to produce merged tokens that capture the semantic content of the cluster, which are subsequently added to the retention pool. Finally, all tokens in the retention pool are rearranged strictly according to their original spatiotemporal ordering to maintain the structural integrity of the image representation.

\begin{figure}[t]
    \centering
    \includegraphics[width=1\linewidth]{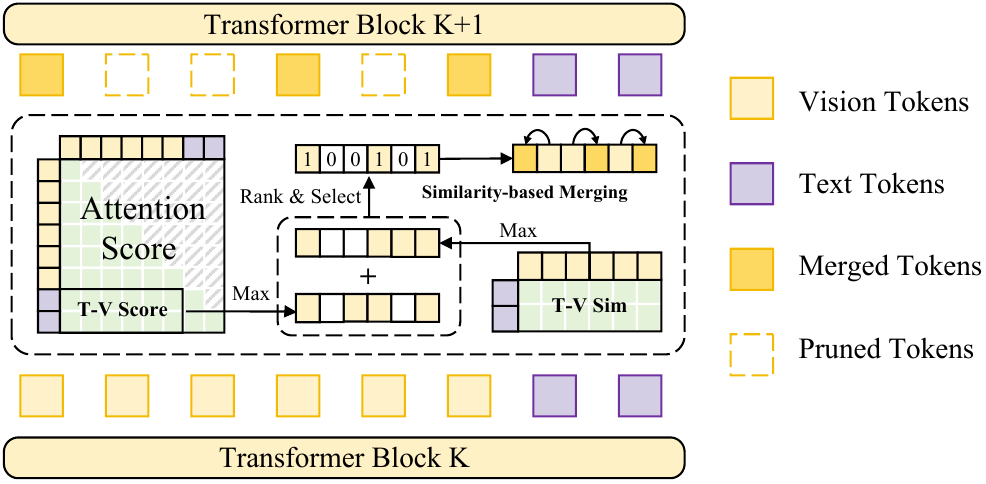}
    \caption{\textbf{The Text-Aware Merging} mechanism further enhances answer accuracy by guiding the LLM to focus on visual tokens that are most relevant to the input query.}
    \label{fig:innerllm}
    \vspace{-10pt}
\end{figure}

\subsection{Text-Aware Merging}

Current inner-LLM pruning methods for Video-LLMs based on last token attention scores often fail to precisely identify visual content closely aligned with the query intent. Moreover, the relative position bias introduced by Rotary Position Embedding (RoPE)~\cite{su2024roformer} leads the attention mechanism to disproportionately focus on spatially adjacent tokens, further compromising the reliability of attention-based token selection.

To overcome these limitations, we introduce the text-aware token merging strategy depicted in~\cref{fig:innerllm}. The approach leverages the attention distribution from text to visual tokens to identify and retain semantically relevant visual content. At the same time, it incorporates a semantic similarity measure between text and visual tokens to reduce positional sensitivity caused by over-reliance on attention weights. This dual mechanism enhances both the accuracy and robustness of token selection and fusion.

Specifically, we construct a decision score $I(v_i)$ for each visual token $v_i$ by integrating cross-attention and cosine similarity metrics. First, we use~\cref{eq:attn} to compute the attention score matrix $A~\in~\mathbb{R}^{(N_v^{\prime} + N_q) \times (N_v^{\prime} + N_q)}$, where $N_v^{\prime}$ and $N_q$ denote the number of visual tokens retained and text tokens, respectively. 
The submatrix $A_{qv}$ is extracted from the matrix $A$ for each visual token with respect to the text tokens. 
We then compute the maximum value $A_m$ across the text dimension and normalize it to obtain $A_m^{\mathrm{norm}}$ as the final attention score for each visual token.
The formulas are as follows:
\begin{equation}
    A_{qv}=A[N_v^{\prime}:,:N_v^{\prime}],
\end{equation}
\begin{equation}
    A_m = \max_{i=1}^{N_q} A_{qv}(i,:), 
\end{equation}
\begin{equation}
    A_m^\mathrm{norm}=\frac{A_m-\min(A_m)}{\max(A_m)-\min(A_m)}.
    \label{eq:norm}
\end{equation}

Next, to quantify the semantic similarity, we compute the maximum cosine similarity between each visual token $v_i$ and all text tokens:
\begin{equation}
    S_m(v_i) = \max_{t \in \mathcal{T}} \frac{v_i \cdot t}{\|v_i\| \|t\|},
\end{equation}
where the set $\mathcal{T}$ represents all text tokens, and $t$ denotes an arbitrary element from $\mathcal{T}$. Normalize $S_m(v_i)$ using the same method as~\cref{eq:norm} to calculate $S^{\mathrm{norm}}_m(v_i)$.

The final decision score $I(v_i)$ is a weighted sum of these two components, balanced by a hyperparameter $\lambda$:
\begin{equation}
    I(v_i) = (1-\lambda) \cdot A_m^\mathrm{norm}(v_i) + \lambda \cdot S^{\mathrm{norm}}_m(v_i).
\end{equation}
Based on the decision scores $I(v_i)$, the visual tokens are partitioned into two mutually exclusive subsets: the retaining set $\mathcal{V}_{\mathrm{retaining}}$, which comprises the top $R\%$of tokens with the highest scores, and the pruning set $\mathcal{V}_{\mathrm{pruning}}$, consisting of the remaining tokens. To fully leverage the information from the pruned tokens, they are not discarded directly. 
Instead, each pruned visual token $v_j \in \mathcal{V}_{\mathrm{pruning}}$ is merged into the most semantically similar retained token within $\mathcal{V}_{\mathrm{retaining}}$ 
based on cosine similarity. Specifically, the index of the most similar retained token is determined as:
\begin{equation}
r(j) = \arg\max_{v_k \in \mathcal{V}_{\text{retaining}}}
\frac{v_j \cdot v_k}{\|v_j\| \, \|v_k\|}.
\end{equation}
Subsequently, $v_j$ is integrated into $v_{r(j)}$ through averaging operation, enabling each merged token $\tilde{v}_{r(j)}$ to aggregate the semantic information of its corresponding pruned token.

\section{Results}
\label{sec:exp}

\subsection{Experimental Settings}

\noindent\textbf{Datasets.}~We evaluate the proposed method on several general video understanding benchmarks: MVBench~\cite{li2024mvbench}, EgoSchema~\cite{mangalam2023egoschema}, MLVU~\cite{zhou2025mlvu}, LongVideoBench~\cite{wu2024longvideobench}, and VideoMME~\cite{fu2025videomme}. These datasets encompass videos of varying durations, with MVBench primarily featuring clips under one minute and the others containing longer videos spanning several minutes. By integrating diverse video lengths and scene complexities, these benchmarks collectively provide a broadly representative evaluation framework, effectively validating the performance and generalization capability of our method across a spectrum of tasks.

\noindent\textbf{Baselines.}~To comprehensively evaluate the performance of the proposed method, we compare it against five training-free baselines in visual token compression. The selected methods include 1) FastV~\cite{chen2024image}, which prunes visual tokens within the LLM; 2) VisionZip~\cite{yang2025visionzip}, which employs spatial compression; 3) LLaVA-Scissor~\cite{sun2025llavascissor}, based on semantic connected component analysis; 4) FastVID~\cite{shen2025fastvid}, utilizing density clustering to perform intra-group fusion; and 5) HoliTom~\cite{shao2025holitom}, which applies dynamic programming for token grouping along with spatiotemporal two-stage pruning and internal merging. For all baselines, experiments use their open-source code.

\noindent\textbf{Computing Cost Estimation.}~To quantitatively assess the computational efficiency improvement achieved by token compression, we adopt floating-point operations (FLOPs) during the prefilling and decoding phases as the evaluation metric, following prior work~\cite{chen2024image,xing2024pyramiddrop,tao2025dycoke,shao2025holitom}.

\noindent\textbf{Implementation Details.}~The main experiments are performed using the LLaVA-OneVision-7B~\cite{li2024llava}. To systematically assess the generalization ability of the proposed approach, we additionally integrate and evaluate it on LLaVA-Video-7B~\cite{zhang2024video}, LLaVA-OneVision-0.5B~\cite{li2024llava} and Qwen2.5-VL-7B~\cite{bai2025qwen2}. All experiments are performed on a single GPU, using either an NVIDIA RTX 4090 or an A100. For video sampling and representation, LLaVA-OneVision samples 32 frames per video. Each frame is processed through the visual encoder and the projection layer, yielding $N_v=196$ visual tokens per frame, while LLaVA-Video samples 64 frames per video, with each frame represented by $N_v=169$ visual tokens. Parameters of the method are set as follows: the token similarity threshold $\tau$ is 0.7, and the clustering ratio is 0.3. During the internal pruning phase in the LLM, the mechanism is activated starting from layer $K=18$, preserving the top $R=50\%$ of visual tokens, and the $\lambda$ is 0.5. All evaluations are performed using LMMs-Eval~\cite{zhang2024lmmsevalrealitycheckevaluation, lmms_eval2024}.

\begin{table*}[t]
\centering
\caption{\textbf{Comparison of State-of-The-Art Methods on LLaVA-OneVision-7B.} The A\%/B\% retention ratio indicates that A\% of the LLM input tokens are retained, and subsequently compressed to B\% during the LLM forward pass. \textbf{Best} results are in bold, \underline{second best} underlined. ``(w/o M)" means our method without inner-LLM merging.}
\vspace{-10pt}
\label{tab:main_results}
\setlength{\tabcolsep}{3pt}
\setlength{\extrarowheight}{-2pt}
\small
\begin{tabular}{l|ccc|ccccc|cc}
\toprule
\multirow{2}{*}{\bf Method} & 
\multirow{2}{*}{\makecell{\bf FLOPs  \\ \bf (T) $\downarrow$}} &
\multirow{2}{*}{\makecell{\bf FLOPs \\ \bf Ratio $\downarrow$}} & 
\multirow{2}{*}{\makecell{\bf Retention \\ \bf Ratio}} & 
\multirow{2}{*}{\makecell{\bf MVBench \\ $\uparrow$}} &
\multirow{2}{*}{\makecell{\bf EgoSchema \\ $\uparrow$}} &
\multirow{2}{*}{\makecell{\bf MLVU \\ $\uparrow$}} &
\multirow{2}{*}{\makecell{\bf LongVideo \\ \bf Bench $\uparrow$}} &
\multirow{2}{*}{\makecell{\bf VideoMME \\ $\uparrow$}} &
\multicolumn{2}{c}{\bf Avg. $\uparrow$} \\
& & &  & & & & & & \bf Score & \bf \%\\
\midrule
\rowcolor{gray!20}
LLaVA-OV-7B~\cite{li2024llava} & 41.4 & 100\% & 100\% & 58.3 & 60.4 & 47.7 & 56.4 & 58.6 & 56.3 & 100 \\
FsatV~\cite{chen2024image} & 7.9 & 19.1\% & 100\%/10\% & 53.2 & 55.9 & 41.6 & 52.1 & 52.7 & 51.1 & 90.8 \\
VisionZip~\cite{yang2025visionzip} & 4.0  & 9.6\% & 10\% & 53.5 & 58.0 & 42.5 & 49.3 & 53.4 & 51.3 & 91.2 \\
LLaVA-Scissor~\cite{sun2025llavascissor} & 4.0 & 9.6\% & 10\% & - & 57.5 & - & - & 55.8 & - & - \\
FastVID~\cite{shen2025fastvid} & 4.0 & 9.6\% & 10\% & 55.9 & 58.7 & 42.6 & 56.3 & \bf 57.3 & 54.2 & 96.2 \\
HoliTom~\cite{shao2025holitom} & 3.4 & 8.2\% & 10\%/5\% & 57.3 & \bf 61.2 & \underline{45.1} & 56.3 & \underline{56.8} & \underline{55.3} & \underline{98.3} \\
\bf Ours (w/o M) & 4.0 & 9.6\% & 10\% & \underline{57.4} & 60.5 & 44.7 & \bf 57.4 & 56.6 & \underline{55.3}  & \underline{98.3} \\
\bf Ours & 3.4 & 8.2\% & 10\%/5\% & \bf 57.7 & \underline{60.6} & \bf 45.5 & \underline{56.4} & 56.6 & \bf 55.4 & \bf 98.4 \\

\midrule

FastV~\cite{chen2024image} & 6.4 & 15.5\% & 100\%/5\% & 51.2 & 53.9 & 35.8 & 47.9 & 49.7 & 47.7 & 84.7 \\
VisionZip~\cite{yang2025visionzip} & 2.2 & 5.4\% & 5\% & 45.2 & 51.9 & 37.4 & 46.4 & 48.2 & 45.8 & 81.4 \\
LLaVA-Scissor~\cite{sun2025llavascissor} & 2.2 & 5.4\% & 5\% & - & 56.6 & - & - & 53.3 & - & - \\
FastVID~\cite{shen2025fastvid} & 2.2 & 5.4\% & 5\% & 53.0 & 57.1 & 42.1 & 51.1 & 54.2 & 51.5 & 91.5 \\
HoliTom~\cite{shao2025holitom} & 2.0 & 4.7\% & 5\%/2.5\% & \underline{55.6} & \bf 60.5 & 40.6 & 53.6 & 54.2 & 52.9 & 94.0 \\
\bf Ours (w/o M) & 2.2 & 5.4\% & 5\% & \bf 56.4 & 59.5 & \underline{42.5} & \underline{53.7} & \bf 55.0 & \underline{53.4} & \underline{94.9} \\
\bf Ours & 2.0 & 4.7\% & 5\%/2.5\% & \bf 56.4 & \underline{60.2} & \bf 42.8 & \bf 54.5 & \underline{54.7} & \bf 53.7 & \bf 95.4 \\

\midrule

FastV~\cite{chen2024image} & 5.5 & 13.3\% & 100\%/2\% & 49.0 & 50.6 & 34.1 & 47.1 & 47.3 & 45.6 & 81.0 \\
VisionZip~\cite{yang2025visionzip} & 1.2 & 2.9\% & 2\% & 41.7 & 47.6 & 31.8 & 45.1 & 45.9 & 42.4 & 75.3\\
FastVID~\cite{shen2025fastvid} & 1.2 & 2.9\% & 2\% & 48.0 & 52.3 & 37.6 & 47.3 & 49.2 & 46.9 & 83.3 \\
HoliTom~\cite{shao2025holitom} & 1.1 & 2.6\% & 2\%/1\%  & 52.6 & \underline{57.2} & 37.4 & 48.5 & 51.1 & 49.4 & 87.7 \\
\bf Ours (w/o M) & 1.2 & 2.9\% & 2\% & \bf 52.9 & \underline{57.2} & \underline{39.5} & \bf 51.0 & \underline{51.3} & \underline{ 50.4 }& \underline{89.5}\\
\bf Ours & 1.1 & 2.6\% & 2\%/1\% & \underline{52.8} & \bf 57.6 & \bf 40.3 & \underline{50.8} & \bf 51.8 & \bf 50.7 & \bf 90.1 \\

\midrule

FastV~\cite{chen2024image} & 5.2 & 12.6\% & 100\%/1\% & 48.2 & 48.8 & 32.3 & 45.5 & 46.2 & 44.2 & 78.5 \\
VisionZip~\cite{yang2025visionzip} & 0.9 & 2.1\% & 1\% & 40.8 & 43.8 & 29.7 & 44.4 & 44.3 &  40.6 & 72.1 \\
FastVID~\cite{shen2025fastvid} & 0.9 & 2.1\% & 1\% & 45.3 & 47.8 & 32.4 & 46.1 & 47.0 & 43.7 & 77.7 \\
HoliTom~\cite{shao2025holitom} & 0.8 & 2.0\% & 1\%/0.5\% & \underline{49.6} &  52.9 & \underline{33.9} & 48.1 & 49.0 & 46.7  & 82.9 \\
\bf Ours (w/o M) & 0.9 & 2.1\% & 1\% & \bf 50.5 & \underline{53.3} & \bf 34.4 & \bf 49.1 & \bf 49.8 & \bf 47.4 & \bf 84.2 \\
\bf Ours & 0.8 & 2.0\% & 1\%/0.5\% & \bf 50.5 & \bf 53.8 & \bf 34.4 & \underline{48.8} & \underline{49.2} & \underline{47.3} & \underline{84.1} \\

\bottomrule
\end{tabular}
\vspace{-10pt}
\end{table*}
\begin{table*}[t]
\centering
\caption{\textbf{Cross-backbone Method Comparison.} Performance comparison of our method against state-of-the-art methods across different backbones, demonstrating consistent effectiveness.}
\vspace{-10pt}
\label{tab:generalizability_results}
\setlength{\tabcolsep}{3pt}
\setlength{\extrarowheight}{-2pt}
\small
\begin{tabular}{l|ccc|ccccc|cc}
\toprule
\multirow{2}{*}{\bf Method} & 
\multirow{2}{*}{\makecell{\bf FLOPs \\ \bf (T) $\downarrow$}} &
\multirow{2}{*}{\makecell{\bf FLOPs \\ \bf Ratio $\downarrow$}} & 
\multirow{2}{*}{\makecell{\bf Retention \\ \bf Ratio}} & 
\multirow{2}{*}{\makecell{\bf MVBench \\ $\uparrow$}} &
\multirow{2}{*}{\makecell{\bf EgoSchema \\ $\uparrow$}} &
\multirow{2}{*}{\makecell{\bf MLVU \\ $\uparrow$}} &
\multirow{2}{*}{\makecell{\bf LongVideo \\ \bf Bench $\uparrow$}} &
\multirow{2}{*}{\makecell{\bf VideoMME \\ $\uparrow$}} &
\multicolumn{2}{c}{\bf Avg. $\uparrow$} \\
& & &  & & & & & & \bf Score & \bf \% \\
\midrule

\rowcolor{gray!20}
LLaVA-Video-7B~\cite{zhang2024video} & 80.9 & 100\% & 100\% & 60.4 & 57.2 & 53.3 & 58.9 & 64.3 & 58.8 & 100 \\

FsatV~\cite{chen2024image} & 10.2 & 12.6\% & 100\%/2\% & 42.9 & 41.0 & 34.3 & 45.6 & 48.1 & 42.4 & 72.1 \\
VisionZip~\cite{yang2025visionzip} & 1.5 & 1.9\% & 2\% & 43.8 & 39.3 & 33.5 & 45.8 & 48.4 & 42.2 & 71.8 \\
FastVid~\cite{shen2025fastvid} & 1.5 & 1.9\% & 2\% & 48.3 & 45.7 & 37.3 & 49.3 & 53.6 & 46.8 & 79.6 \\
HoliTom~\cite{shao2025holitom} & 1.4 & 1.7\% & 2\%/1\% & \textbf{50.2} & 46.5 & \underline{39.9} & \textbf{50.7} & 55.3 & 48.5 & 82.5 \\
\textbf{Ours (w/o M)} & 1.5 & 1.9\% & 2\% & 49.8 & \underline{46.7} & \textbf{40.8} & 49.8 & \underline{55.8} & \underline{48.6} & \underline{82.7} \\
\textbf{Ours} & 1.4 & 1.7\% & 2\%/1\% & \underline{50.1} & \textbf{46.8} & \bf{40.8} & \underline{50.2} & \textbf{56.2} & \textbf{48.8} & \textbf{83.0} \\

\midrule
FsatV~\cite{chen2024image} & 9.7 & 11.9\% & 100\%/1\% & 42.2 & 38.8 & 31.1 & 45.0 & 46.0 & 40.6 & 69.1 \\

VisionZip~\cite{yang2025visionzip} & 1.2 & 1.5\% & 1\% & 42.3 & 36.8 & 30.9 & 45.6 & 47.0 & 40.5 & 68.9 \\
FastVid~\cite{shen2025fastvid} & 1.2 & 1.5\% & 1\% & 43.6 & 38.8 & 30.1 & 46.2 & 49.3 & 41.6 & 70.7 \\
HoliTom~\cite{shao2025holitom} & 1.1 & 1.4\% & 1\%/0.5\% & 46.4 & \underline{41.1} & 38.9 & 48.8 & 51.7 & 45.4 & 77.2 \\
\bf Ours (w/o M) & 1.2 & 1.5\% & 1\% & \underline{47.9} & \textbf{44.8} & \underline{40.0} & \underline{48.9} & \underline{54.6} & \underline{47.2} & \underline{80.3} \\
\bf Ours & 1.1 & 1.4\% & 1\%/0.5\% & \textbf{48.0} & \textbf{44.8} & \textbf{40.1} & \textbf{49.4} & \textbf{54.9} & \textbf{47.4} & \textbf{80.6} \\

\midrule

\rowcolor{gray!20}
LLaVA-OV-0.5B~\cite{li2024llava} & 3.54 & 100\% & 100\% & 46.6 & 26.6 & 33.5 & 47.5 & 43.7 & 39.6 & 100 \\

FastV~\cite{chen2024image} & 0.50 & 14.1\% & 100\%/2\% & 39.2 & 21.2 & 22.5 & 38.8 & 34.0 & 31.1 & 78.5 \\
VisionZip~\cite{yang2025visionzip} & 0.07 & 1.9\% & 2\% & 36.3 & 20.0 & 22.6 & 38.4 & 34.6 & 30.4 & 76.8 \\
FastVid~\cite{shen2025fastvid} & 0.07 & 1.9\% & 2\% & 37.9 & 22.9 & 23.5 & 40.2 & 36.9 & 32.3 & 81.6 \\
HoliTom~\cite{shao2025holitom} & 0.06 & 1.8\% & 2\%/1\% & 42.6 & 24.5 & \underline{26.1} & 42.4 & 39.4 & \underline{35.0} & \underline{88.4} \\
\textbf{Ours (w/o M)} & 0.07 & 1.9\% & 2\% & \textbf{43.0} & \underline{24.6} & \textbf{26.4} & \textbf{45.3} & \textbf{40.5} & \textbf{35.9} & \textbf{90.7} \\
\textbf{Ours} & 0.06 & 1.8\% & 2\%/1\% & \underline{42.7} & \textbf{24.8} & \textbf{26.4} & \underline{45.2} & \underline{40.4} & \textbf{35.9} & \textbf{90.7} \\

\midrule
FastV~\cite{chen2024image} & 0.48 & 13.7\% & 100\%/1\% & 35.6 & 19.3 & 21.5 & 37.9 & 32.2 & 29.3 & 74.0\\
VisionZip~\cite{yang2025visionzip} & 0.05 & 1.3\% & 1\% & 35.3 & 19.1 & 19.1 & 39.0 & 33.3 & 29.2 & 73.7 \\
FastVid~\cite{shen2025fastvid} & 0.05 & 1.3\% & 1\% & 36.2 & 20.8 & 20.8 & 40.6 & 35.0 & 30.7 & 77.5 \\
HoliTom~\cite{shao2025holitom} & 0.05 & 1.3\% & 1\%/0.5\% & 41.5 & 23.0 & \bf 25.1 & 42.5 & 37.5 & 33.9 & 85.6 \\
\bf Ours (w/o M) & 0.05 & 1.3\% & 1\% & \underline{41.8} & \underline{23.2} & \underline{23.2} & \underline{43.4} & \underline{38.2} & \underline{34.0} & \underline{85.9} \\
\bf Ours & 0.05 & 1.3\% & 1\%/0.5\% & \bf 42.2 & \bf 24.2 & 23.1 & \bf 44.1 & \bf 40.6 & \bf 34.8 & \bf 87.9 \\

\bottomrule
\end{tabular}
\vspace{-10pt}
\end{table*}

\subsection{Main Results}

\begin{figure}[t]
    \centering

    \begin{subfigure}[b]{1\linewidth}
        \centering
        \includegraphics[width=\linewidth]{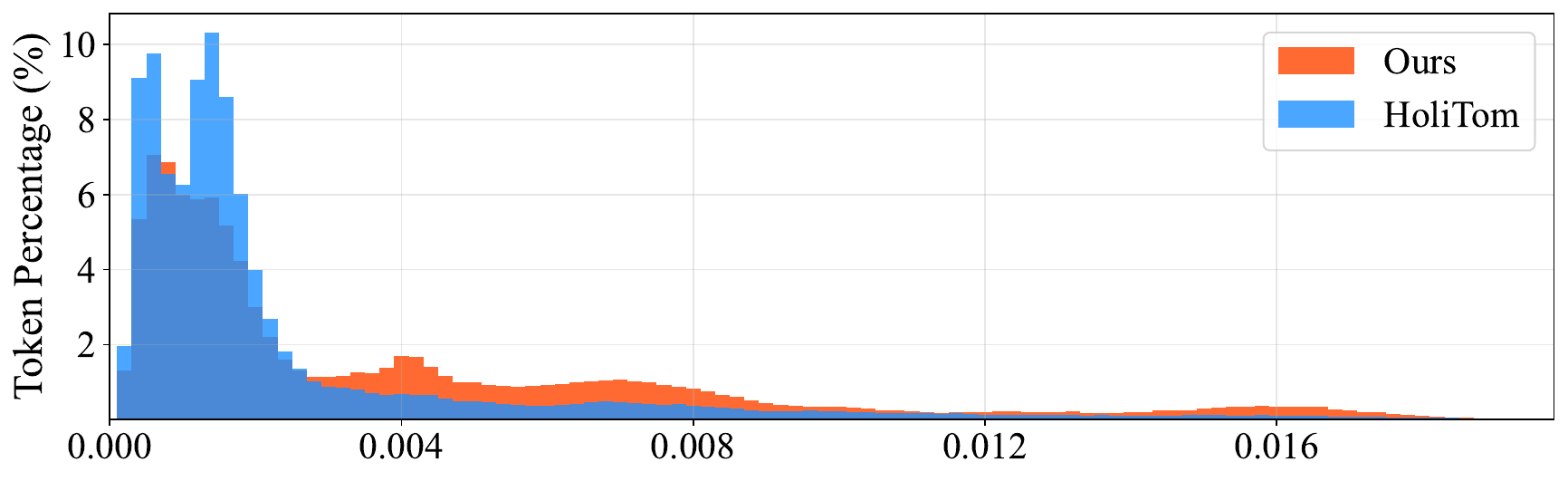}
        \caption{Attention Distribution}
        \label{fig:attn}
    \end{subfigure}
    \begin{subfigure}[b]{1\linewidth}
        \centering
        \includegraphics[width=\linewidth]{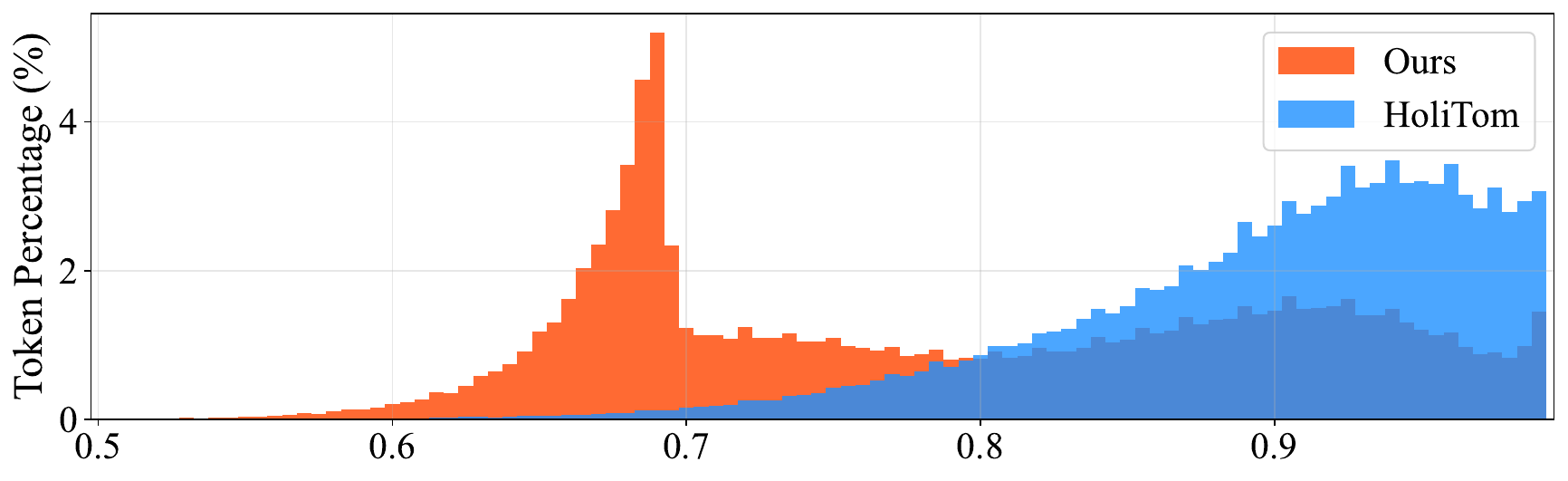}
        \caption{Semantic Similarity Distribution}
        \label{fig:sim}
    \end{subfigure}
    
    \caption{\textbf{Comparative Analysis of Attention Distribution and Semantic Similarity Distribution.} Compared to the HoliTom, our approach more effectively preserves visual tokens with high attention scores but low semantic relevance, thereby eliminating semantic redundancy while retaining visually informative tokens.}
    \label{fig:attn_sim}
    \vspace{-10pt}
\end{figure}

\noindent\textbf{State-of-the-Art Performance.}~As presented in~\cref{tab:main_results}, experimental results demonstrate that our approach consistently achieves state-of-the-art performance under all settings. At a 10\% retention ratio, model performance shows only a slight decrease, maintaining approximately 98.4\% of its original capability. To further examine performance at low retention ratios, we reduce the ratio to 5\%. At this level, FastV~\cite{chen2024image} and VisionZip~\cite{yang2025visionzip} exhibit performance drops of about 15.3\% and 18.6\%, respectively, due to their limited ability to compress temporal redundancy in videos. In contrast, our method preserves 94.9\% of its original performance. With the proposed unified spatiotemporal token pruning strategy, our method maintains about 89.5\% of the baseline performance even when the retention ratio decreases to 2\%. Under extreme compression with the 1\% retention ratio, it still retains 84.2\% of the original performance while reducing the inference computational cost to only 2.1\%. By comparison, FastVid~\cite{shen2025fastvid} maintains 77.7\% of the original performance only at the 1\% retention ratio. Moreover, the proposed text-aware inner-LLM merging strategy for LLMs further improves inference efficiency and reduces computational cost. At the 2\% retention ratio, this strategy achieves about 90.1\% of the original model performance while requiring only 2.6\% of the original FLOPs. As presented in~\cref{fig:attn_sim}, compared to HoliTom, the tokens selected by our method achieve higher attention scores while maintaining lower internal similarity. In addition to the multiple-choice benchmarks, we further validate our method on the open-ended ActivityNet-QA~\cite{yu2019activitynet} benchmark, where it also achieves state-of-the-art performance. Comprehensive results are provided in the supplementary material. This demonstrates its efficacy in eliminating redundancy among highly responsive tokens and suppressing semantic repetition. Consequently, this strategy enables a limited token set to preserve richer and more diverse semantic information.

\noindent\textbf{Different Backbones.}~Our method, designed as a plug-and-play module, demonstrates strong cross-backbone transferability. To validate its versatility and effectiveness, we conduct experiments on LLaVA-Video-7B~\cite{zhang2024video} and LLaVA-OneVision-0.5B~\cite{li2024llava}. The results are presented in~\cref{tab:generalizability_results}. On the LLaVA-OneVision-0.5B model, our method preserves 90.7\% of the original performance while retaining only 2\% of visual tokens. For the more structurally complex and harder-to-compress LLaVA-Video-7B, our method maintains about 80.6\% of the original performance under extreme compression with only 1\% of the tokens retained, surpassing HoliTom, which achieves 77.2\% under the same conditions. Our method was also validated on Qwen2.5-VL-7B~\cite{bai2025qwen2}, with results presented in the supplementary materials. Overall, our approach exhibits strong adaptability across backbone networks and consistently outperforms existing methods even under ultra-low token retention ratios.

\begin{table}[t]
\centering
\caption{Performance of LLaVA-OneVision-7B under different frame sampling rates at 2\% token retention.}
\vspace{-10pt}
\resizebox{\linewidth}{!}{
\begin{tabular}{l|c|c|cccc}
\toprule
\multirow{2}{*}{\textbf{Method}} & \multirow{2}{*}{\textbf{Frames}} & \multirow{2}{*}{\makecell[c]{\textbf{FLOPs}\\\textbf{(T)}}} & \multicolumn{4}{c}{\textbf{VideoMME}} \\
\cline{4-7}
 &  &  & \textbf{Short} & \textbf{Medium} & \textbf{Long} & \textbf{Total} \\
\midrule
\rowcolor{gray!20}
Vanilla~\cite{li2024llava} & 16  & 19.0 & 68.2 & 53.8 & 48.2 & 56.7 \\
HoliTom~\cite{shao2025holitom} & 64  & 1.7  & 63.0 & 53.9 & 46.3 & 54.4 \\
\text{Ours} & 64  & 1.7  & 62.4 & 53.3 & 47.6 & 54.5 \\
HoliTom~\cite{shao2025holitom} & 96  & 2.2  & 64.1 & 52.4 & 45.9 & 54.1 \\
\text{Ours} & 96  & 2.2  & 65.8 & 55.0 & 46.8 & 55.9 \\
HoliTom~\cite{shao2025holitom} & 128 & 2.8  & 66.2 & 53.2 & 46.8 & 55.4 \\
\text{Ours} & 128 & 2.8  & \textbf{68.4} & \textbf{55.3} & \textbf{50.2} & \textbf{58.0} \\
\bottomrule
\end{tabular}
}
\vspace{-10pt}
\label{tab:more_frames}
\end{table}

\noindent\textbf{Higher Frame Sampling Rates.}~We conduct experiments on the LLaVA-OneVision-7B using higher frame sampling rates. As illustrated in~\cref{tab:more_frames}, when the frame count increases to 128 frames, our method achieves a computational complexity of 2.8T FLOPs while retaining only 2\% of tokens. This performance surpasses that of LLaVA-OneVision-7B at the 16-frame setting. Detailed experimental results are listed in the supplementary materials.  These results underscore the efficacy of our method in extending input frame length through an extremely low token retention ratio, thereby significantly enhancing the model's capacity to comprehend long video content.

\begin{table}[t]
\centering
\caption{Performance comparison of LLaVA-OneVision-7B with different methods at high token retention ratios.}
\vspace{-10pt}
\resizebox{\linewidth}{!}{
\begin{tabular}{l|c|cc|c}
\toprule
\textbf{Method} &
\makecell[c]{\textbf{Retention}\\\textbf{Ratio}} &
\makecell[c]{\textbf{MVBench}\\} &
\makecell[c]{\textbf{LongVideo}\\\textbf{Bench}} &
\makecell[c]{\textbf{Avg. Score (\%)}} \\
\midrule
\rowcolor{gray!10}
Vanilla~\cite{li2024llava}   & 100\% & 58.3 & 56.4 & 57.4 (100.0\%) \\
FastVid~\cite{shen2025fastvid}   & 25\%  & 56.5 & 56.3 & 56.4 (98.3\%) \\
VisionZip~\cite{yang2025visionzip} & 25\%  & 57.9 & 56.5 & 57.2 (99.7\%) \\
HoliTom~\cite{shao2025holitom}   & 15\%  & 58.1 & 56.4 & 57.3 (99.8\%) \\
\textbf{Ours} & 15\%  & \textbf{58.3} & \textbf{57.5} & \textbf{57.9 (100.9\%)} \\
\bottomrule
\end{tabular}
}
\vspace{-10pt}
\label{tab:moreratio}
\end{table}
\noindent\textbf{Higher Token Retention Ratios.}~According to \cref{tab:moreratio}, our method maintains superior performance at retention ratios above 10\%, yet a performance plateau is observed for all methods in this range. We therefore focus our evaluation on more challenging scenarios with lower retention ratios. Detailed experimental results are included in the supplementary materials.

\noindent\textbf{Computational Efficiency.}~As reported in~\cref{tab:ttft_summary}, our method delivers substantial efficiency gains. Although our TTFT is slightly higher than that of FastVID, this is compensated by a superior throughput and performance. Consequently, our approach surpasses the current SOTA method, HoliTom, in overall efficiency.

\begin{table*}[t]
\centering
\caption{Module Ablations for Unified Spatiotemporal Token Compression.}
\vspace{-10pt}
\label{tab:ablations}
\setlength{\extrarowheight}{-2pt}
\begin{tabular}{ccc|ccccc|cc}

\toprule

\multirow{2}{*}{\makecell{\bf Attn \\}} &
\multirow{2}{*}{\makecell{\bf Sim \\}} &
\multirow{2}{*}{\makecell{\bf Cluster \\}} &
\multirow{2}{*}{\makecell{\bf MVBench \\ }} &
\multirow{2}{*}{\makecell{\bf EgoSchema \\ }} &
\multirow{2}{*}{\makecell{\bf MLVU \\ }} &
\multirow{2}{*}{\makecell{\bf LongVideo \\ \bf Bench }} &
\multirow{2}{*}{\makecell{\bf VideoMME \\ }} &
\multicolumn{2}{c}{\bf Avg. $\uparrow$} \\
& & &  & & & & & \bf Score & \bf \%\\
\midrule

\checkmark & & & 41.4 & 46.6 & 30.8 & 44.7 & 45.3 & 41.8 & 74.2 \\
& \checkmark & & 51.4 & 55.1 & 38.8 & 50.0 & 50.9 & 49.2 & 87.5 \\
& & \checkmark & 52.2 & 55.0 & 36.3 & 48.4 & 51.3 & 48.6 & 86.4 \\

\checkmark & \checkmark & & 52.1 & 56.8 & 38.4 & 49.9 & 51.2 & 49.7 & 88.2 \\
\checkmark & & \checkmark & 48.3 & 51.4 & 33.2 & 46.4 & 49.5 & 45.8 & 81.3 \\
& \checkmark & \checkmark & 51.8 & 55.5 & 38.4 & 50.6 & 51.0 & 49.5 & 87.9 \\
\checkmark & \checkmark & \checkmark  & \textbf{52.9} & \textbf{57.2} & \textbf{39.5} & \textbf{51.0} & \textbf{51.3} & \textbf{50.4} & \textbf{89.5} \\
\bottomrule
\end{tabular}
\vspace{-10pt}
\end{table*}
\begin{figure*}[t]
    \centering 
    \begin{subfigure}[b]{0.33\linewidth}
        \centering
        \includegraphics[width=1\linewidth]{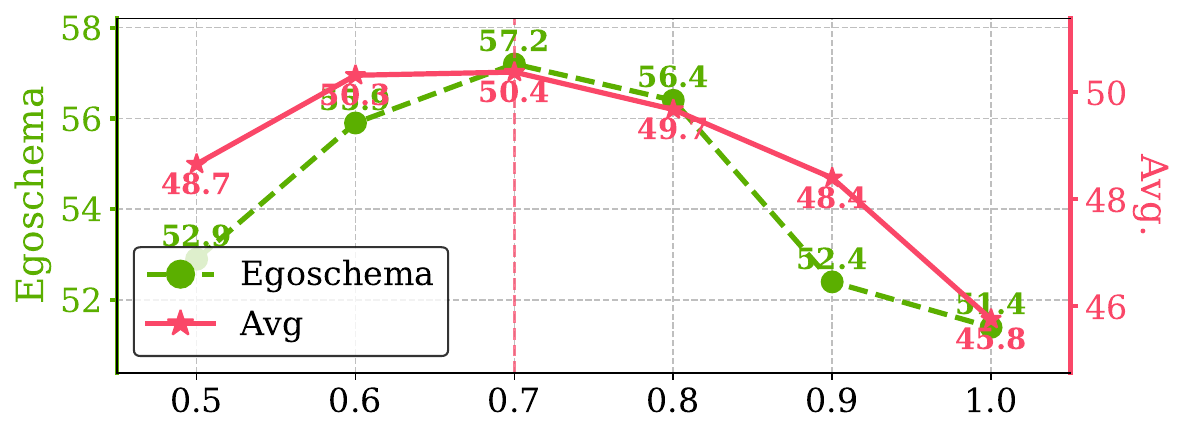} 
        \caption{Ablation on $\tau$.}
        \label{fig:tau}
    \end{subfigure}
    \hfill 
    \begin{subfigure}[b]{0.33\linewidth}
        \centering
        \includegraphics[width=1\linewidth]{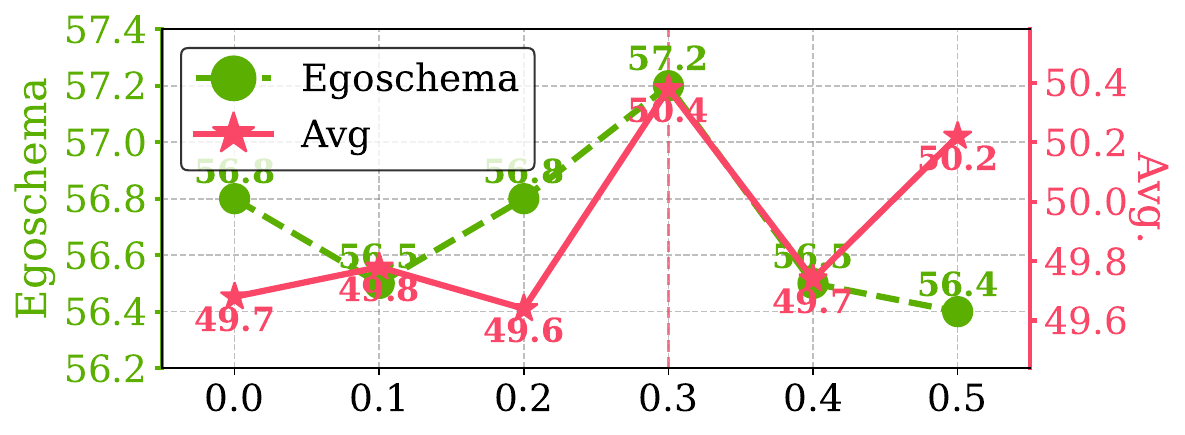} 
        \caption{Ablation on Clustering Token Ratio.}
        \label{fig:dpc}
    \end{subfigure}
    \hfill 
    \begin{subfigure}[b]{0.33\linewidth}
        \centering
        \includegraphics[width=1\linewidth]{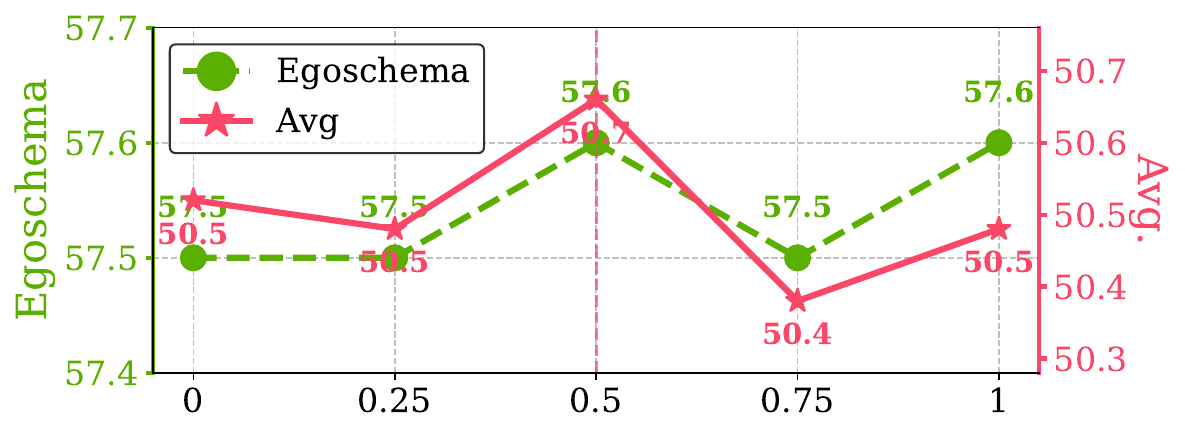} 
        \caption{Ablation on $\lambda$.}
        \label{fig:lambda}
    \end{subfigure}
    \caption{Ablation Experiment Results for Each Parameter.} 
    \label{fig:abla} 
    \vspace{-10pt}
\end{figure*}

\begin{table}[t]
\small
\caption{
\textbf{Comparison of Efficiency and Performance.} Process, Prefill, and Time to First Token (TTFT) are measured in milliseconds; throughput is reported in tokens per second (token/s).
}
\vspace{-10pt}
\centering
\renewcommand{\arraystretch}{1.1}
\resizebox{\columnwidth}{!}{
\begin{tabular}{l|c|c|c|c|c}
\toprule
\makecell[c]{\textbf{Method}} &
\makecell[c]{\textbf{Process} }&
\makecell[c]{\textbf{Prefill}} &
\makecell[c]{\textbf{TTFT}} &
\makecell[c]{\textbf{Throughput}} &
\makecell[c]{\textbf{Score}}\\
\midrule
\rowcolor{gray!20}
Vanilla~\cite{li2024llava}    & - & 701.2 (1×)  & 1104.7 & 27.6 & 56.3 \\
VisionZip~\cite{yang2025visionzip} & 35.8 & 42.1 (17×)  & 451.6  & 35.4 & 45.8 \\
FastVid~\cite{shen2025fastvid}   & \bf 8.6 & 43.1 (16×)  & \bf 446.2  & 33.8 & 46.9 \\
HoliTom~\cite{shao2025holitom}   & 88.7     & 40.5 (17×)  & 497.4  & 34.3 & 49.4 \\
Ours      & 74.0  & \bf 31.3 (22×)  & 473.8  & \bf 35.7 & \bf 50.7 \\
\bottomrule
\end{tabular}
}
\vspace{-10pt}
\label{tab:ttft_summary}
\end{table}

\subsection{Ablation Studies}
\noindent\textbf{Attention-based Selection.}~Our ablation study confirms the necessity of the attention-based selection mechanism. As presented in~\cref{tab:ablations}, the results show that using token similarity as the sole criterion for redundancy removal leads to a notable performance drop, which is particularly evident at the 2\% token retention ratio, where performance retains only 87.5\%.

\noindent\textbf{Similarity-based Pruning.}~To evaluate the effectiveness of the similarity-based pruning mechanism, we conduct ablation experiments at a 2\% token retention ratio. In this setting, tokens are selected solely according to high attention scores without applying similarity-based pruning. Based on~\cref{tab:ablations}, incorporating similarity pruning improves model performance by about 14\%. Furthermore, a parameter ablation on the pruning threshold $\tau$ reveals that performance peaks at $\tau=0.7$, demonstrated in~\cref{fig:tau}, achieving the highest performance retention rate of 89.5\%.

\noindent\textbf{Clustering-based Merging.}~To evaluate the contribution of the clustering-based merging strategy, we conduct ablation experiments on this module. As shown in~\cref{tab:ablations}, introducing the clustering-based token merging method improves model performance by about 1.3\%. By consolidating semantically similar tokens within the discard pool, this approach effectively preserves comprehensive semantic information across the video sequence. A subsequent parameter study on the cluster token proportion reveals an optimum at 0.3, illustrated in~\cref{fig:dpc}.

\begin{figure}[t]
    \centering
    \includegraphics[width=1\linewidth]{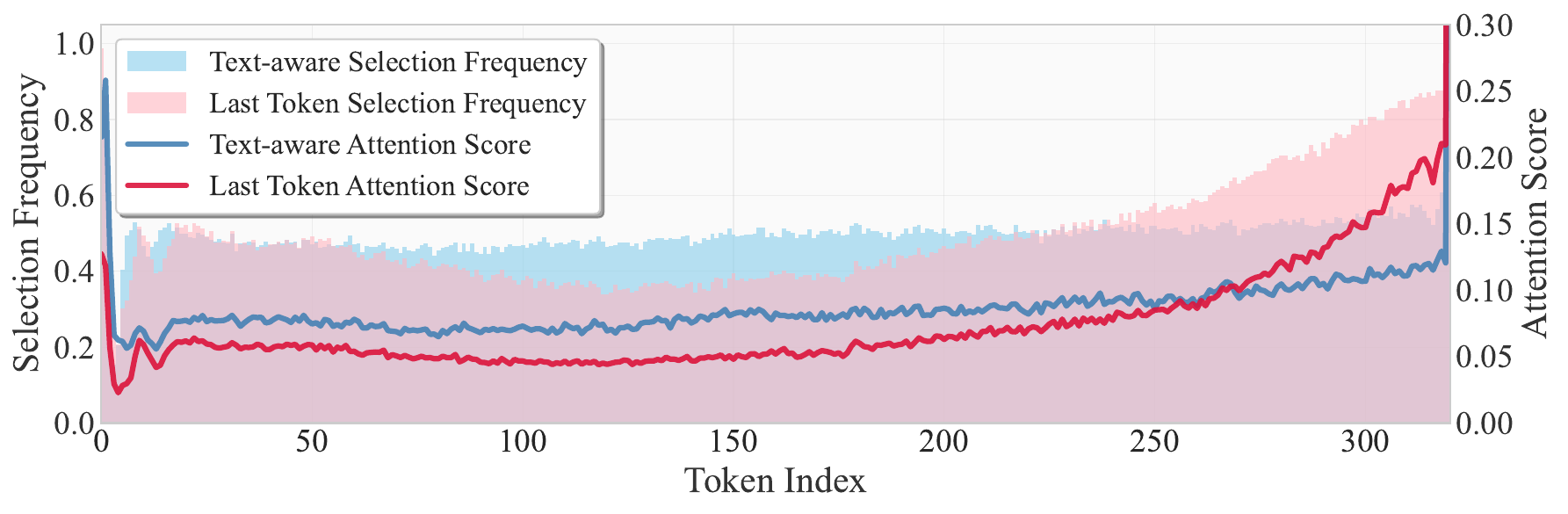}
    \caption{Distributions of Selected Token Indices for Text-Aware and Last Token Strategies on the MVBench.}
    \label{fig:llm_attn}
    \vspace{-10pt}
\end{figure}

\noindent\textbf{Text-Aware Merging.}~We conduct ablation studies to evaluate the text-aware inner-LLM merging mechanism. Our results demonstrate that relying exclusively on either attention scores or cosine similarity leads to a notable drop in performance retention. In contrast, combining both factors produces the best overall performance. As illustrated in~\cref{fig:lambda}, model performance reaches its peak when the $\lambda$ is set to 0.5. Moreover, findings from~\cref{fig:llm_attn} indicate that using attention scores from the last token introduces positional bias by overemphasizing tokens near the end of the sequence. By comparison, our proposed text-aware strategy alleviates this issue and promotes a more balanced probability distribution across all token positions, improving the average performance retention rate from 93.8\% to 95.4\% at a 5\% token retention ratio.
 
\section{Conclusion}
\label{sec:conclusion}

We propose a unified spatiotemporal and text-aware framework for ultra-low-ratio visual token compression in Video-LLMs. By combining attention-based contribution scores with semantic similarity, our pruning strategy selects informative and non-redundant visual tokens. The text-aware merging mechanism further preserves tokens most relevant to textual prompts, ensuring strong semantic alignment under extreme compression. 
Our method requires no training, supports plug-and-play use, and generalizes well across Video-LLMs. We believe this work points the way toward scalable deployment of Video-LLMs, enabling efficient real-world applications without sacrificing accuracy.

\section{Acknowledgement}
This work was supported by the National Key Re-search and Development Program of China under Grant 2024YFF0509700, National Natural Science Foundation of China (62471290, 62431015, 62331014) and the Fundamental Research Funds for the Central Universities.

{
    \small
    \bibliographystyle{ieeenat_fullname}
    \bibliography{main}
}

\clearpage
\setcounter{page}{1}
\maketitlesupplementary

\renewcommand{\thefigure}{S\arabic{figure}}  
\renewcommand{\theequation}{S\arabic{equation}} 
\renewcommand{\thetable}{S\arabic{table}}

\setcounter{table}{0}
\setcounter{figure}{0}
\setcounter{equation}{0}

\section{Preliminaries}

The inference process of Video-LLMs typically involves three phases: \textit{encoding}, \textit{prefilling}, and \textit{decoding}.

\noindent\textbf{(1) Encoding Phase.} Given an input video containing $B$ frames, the visual encoder first processes each frame individually to generate $N_v$ visual embedding vectors. A projector then maps these visual embeddings into the text embedding space, producing the visual tokens $\boldsymbol{H}_v \in \mathbb{R}^{BN_v \times d}$, where $d$ denotes the dimension of the LLM hidden state. 
Simultaneously, the text prompt $T=\{t_i\}_{i=1}^{N_q}$ is tokenized and passed through an embedding layer to obtain the token embeddings $\boldsymbol{H}_q \in \mathbb{R}^{N_q \times d}$. 
Finally, the visual tokens $\boldsymbol{H}_v$ and text token embeddings $\boldsymbol{H}_q$ are concatenated into a unified sequence $\boldsymbol{H} = \mathrm{concat}[\boldsymbol{H}_v, \boldsymbol{H}_q]$, which serves as the input to the LLM.

\noindent\textbf{(2) Prefilling Phase.} During the prefilling phase, each Transformer layer $l$ in the LLM conducts self-attention over the input $\boldsymbol{H}$. Specifically, each layer first projects $\boldsymbol{H}$ into query $\boldsymbol{Q}^l$, key $\boldsymbol{K}^l$, and value $\boldsymbol{V}^l$ matrices, calculated according to the following formula:
\begin{equation}
    \boldsymbol{Q}^l = \boldsymbol{H}\boldsymbol{W}_Q^l, \quad \boldsymbol{K}^l = \boldsymbol{H}\boldsymbol{W}_K^l, \quad \boldsymbol{V}^l = \boldsymbol{H}\boldsymbol{W}_V^l, 
\end{equation}
where $\boldsymbol{W}_Q^l, \boldsymbol{W}_K^l, \boldsymbol{W}_V^l \in \mathbb{R}^{d \times d}$ are learnable projection matrices.
The resulting $\boldsymbol{K}^l$ and $\boldsymbol{V}^l$ matrices are then stored in the KV cache to expedite computation during the subsequent decoding phase.

\noindent\textbf{(3) Decoding Phase.} During the decoding phase, the model generates output tokens autoregressively while dynamically accessing and updating the KV cache. At each time step $t$, the LLM computes the query, key, and value solely from the latest generated token $h_t$. The resulting $\boldsymbol{K}$ and $\boldsymbol{V}$ are then appended incrementally to the KV cache:
\begin{equation}
    \boldsymbol{K} = [\boldsymbol{K}, h_t\boldsymbol{W}_K], \quad \boldsymbol{V} = [\boldsymbol{V}, h_t\boldsymbol{W}_V],
\end{equation}
This process eliminates redundant attention computations over previously processed tokens and substantially enhances decoding efficiency.

\section{BenchMarks}
\label{sec:benchmarks}
We evaluate our method on various video understanding benchmarks, detailed as follows:

\begin{itemize}
    \item \textbf{MVBench}~\cite{li2024mvbench} formulates 20 video understanding tasks, each with 200 QA pairs, to assess temporal comprehension beyond single-frame analysis and provide a comprehensive model evaluation.

    \item 
    \textbf{EgoSchema}~\cite{mangalam2023egoschema}
    consists of over 5000 human-curated multiple-choice question answer pairs, spanning over 250 hours of real video data, covering a very broad range of natural human activity and behavior. For each question, EgoSchema requires the correct answer to be selected between five given options based on a three-minute-long video clip.

     \item \textbf{MLVU}~\cite{zhou2025mlvu} designed for long-form videos, features a benchmark with durations from 3 minutes to over 2 hours (12-minute average). It covers diverse genres like movies, documentaries, and TV series, while evaluating models across 9 distinct tasks such as topic reasoning, video summarization, and needle question answering.
     
    \item \textbf{LongVideoBench}~\cite{wu2024longvideobench} consists of 3,763 videos and 6,678 associated multiple-choice questions, covering diverse domains such as movies and news, specifically designed to assess a model’s capacity for temporal information retrieval and analysis.

    \item \textbf{VideoMME}~\cite{fu2025videomme} consists of 900 videos and 2,700 QA pairs with durations from 11 seconds to 1 hour. These are categorized into three temporal subsets (short-, medium-, and long-term) and span 6 main visual domains, such as life record and knowledge.

    \item \textbf{ActivityNet-QA}~\cite{yu2019activitynet} contains 58,000 human-annotated QA pairs on 5,800 videos derived from the popular ActivityNet dataset. It provides a benchmark for testing the performance of VideoQA models on long-term spatiotemporal reasoning. Moreover, LLMs are employed to assess the quality of model-generated responses, providing a more flexible and nuanced evaluation compared to traditional accuracy-based metrics.

\end{itemize}

\begin{table*}[t!]

\caption{Comparison results based on the Qwen2.5-VL-7B model.}
\centering

\begin{tabular}{l| c |c c c c c| c c}
\toprule
\multirow{2}{*}{\bf Method} &
\multirow{2}{*}{\makecell{\bf Retention\\\bf Ratio}} &
\multirow{2}{*}{\makecell{\bf MVBench \\ }} &
\multirow{2}{*}{\makecell{\bf EgoSchema \\ }} &
\multirow{2}{*}{\makecell{\bf MLVU \\ }} &
\multirow{2}{*}{\makecell{\bf LongVideo \\ \bf Bench }} &
\multirow{2}{*}{\makecell{\bf VideoMME \\ }} &
\multicolumn{2}{c}{\bf Avg. $\uparrow$} \\
& & & & & & & \bf Score & \bf \% \\
\midrule
\rowcolor{gray!20}
Qwen2.5-VL~\cite{bai2025qwen2} & 100\% & 
63.0 & 52.8 & 38.5 & 55.0 & 57.5 & 53.4 & 100.0 \\
\midrule
VisionZip~\cite{yang2025visionzip}          & 2\% & 
50.3 & 45.3 & 29.7 & 43.4 & 44.1 & 42.6 & 79.8 \\
FastVID~\cite{shen2025fastvid}            & 2\% & 
51.8 & 46.2 & 30.6 & 45.0 & \underline{46.4} & 44.0 & 82.5 \\
\bf Ours(w/o M)               & 2\% & 
\textbf{54.1} & \underline{47.8} & \textbf{32.7} & \underline{45.4} & \textbf{48.1} & \textbf{45.6} & \textbf{85.5} \\
\bf Ours            & 2\% & 
\underline{53.7} & \textbf{48.4} & \underline{31.8} & \textbf{45.6} & \textbf{48.1} & \underline{45.5} & \underline{85.3} \\
\midrule
VisionZip~\cite{yang2025visionzip}& 1\% & 
46.1 & 42.3 & 25.8 & 41.8 & 41.4 & 39.5 & 74.0 \\
FastVID~\cite{shen2025fastvid}& 1\% & 
47.2 & 43.5 & 26.4 & \textbf{45.1} & 43.1 & 41.1 & 76.9 \\
\bf Ours(w/o M)               & 1\% & 
\underline{50.9} & \underline{45.0} & \underline{28.3} & 43.2 & \underline{44.4} & \underline{42.4} & \underline{79.4} \\
\bf Ours            & 1\% & 
\textbf{51.0} & \textbf{45.5} & \textbf{28.6} & \underline{43.4} & \textbf{44.5} & \textbf{42.6} & \textbf{79.8} \\
\bottomrule
\end{tabular}
\label{tab:qwen}
\end{table*}

\section{Computing Cost Estimation.}
To quantitatively assess the computational efficiency improvement achieved by token compression, we adopt floating-point operations (FLOPs) during the prefilling and decoding phases as the evaluation metric, following prior work~\cite{chen2024image,xing2024pyramiddrop,tao2025dycoke,shao2025holitom}. The total FLOPs of the Transformer model are formulated as:
\begin{multline}
\label{eq:flops}
\text{FLOPs} = \sum_{i=1}^{T}((4n_id^2 + 2n_{i}^2d + 2n_idm) + \\
R((4d^2 + 2dm) + 2(dn_i + \frac{d}{2}(R+1))),
\end{multline}
where $T$ denotes the total number of layers, $n_i$ indicates the number of input tokens after pruning at layer $i$, $d$ is the hidden dimension, $m$ is the intermediate size of the feedforward network (FFN), and $R=100$ is the fixed number of tokens generated during decoding.
\section{Additional Results}
To provide a more comprehensive evaluation, we further include results at a higher retention ratio (15\%) and reproduce DyToK~\cite{lilessfull} under the same setting (\cref{tab:extra_results}). The results show that our method remains state-of-the-art across all benchmarks. Notably, at 15\% retention, our performance nearly matches that of the full model. Meanwhile, our advantage is more pronounced in low-retention regimes, where efficient token utilization becomes critical. As the retention ratio increases, all methods gradually converge to a similar performance plateau.

\begin{table*}[t]
\centering
\caption{Additional Comparison with DyToK under Higher Retention Ratios}
\label{tab:extra_results}

\begin{tabular}{l| c |c c c c c| c c}
\toprule
\multirow{2}{*}{\bf Method} &
\multirow{2}{*}{\makecell{\bf Retention\\\bf Ratio}} &
\multirow{2}{*}{\makecell{\bf MVBench \\ }} &
\multirow{2}{*}{\makecell{\bf EgoSchema \\ }} &
\multirow{2}{*}{\makecell{\bf MLVU \\ }} &
\multirow{2}{*}{\makecell{\bf LongVideo \\ \bf Bench }} &
\multirow{2}{*}{\makecell{\bf VideoMME \\ }} &
\multicolumn{2}{c}{\bf Avg. $\uparrow$} \\
& & & & & & & \bf Score & \bf \% \\

\midrule
\rowcolor{gray!20}
LLaVA-OV-7B~\cite{li2024llava} & 100\% & 58.3 & 60.4 & 47.7 & 56.4 & 58.6 & 56.3 & 100 \\

VisionZip~\cite{yang2025visionzip} & 15\% & 56.5 & 59.8 & 43.3 & 54.4 & 56.1 & 54.0 & 95.9 \\
DyToK~\cite{lilessfull} & 15\% & 56.1 & 59.5 & 44.6 & 53.7 & 56.1 & 54.0 & 95.9 \\
FastVID~\cite{shen2025fastvid} & 15\% & 56.0 & 58.8 & 43.3 & 56.2 & \bf 57.7 & 54.4 & 96.6 \\
HoliTom~\cite{shao2025holitom} & 15\%/7.5\% & 58.1 & \bf 61.2 & \bf 46.7 & 56.4 & 57.3  & 56.0 & 99.5 \\

\bf Ours & 15\%/7.5\% & \bf 58.3 & {60.5} & 46.5 & \bf{57.5} & \bf 57.7 & \bf 56.1 & \bf 99.6 \\

\midrule
FsatV~\cite{chen2024image}& 100\%/10\% & 53.2 & 55.9 & 41.6 & 52.1 & 52.7 & 51.1 & 90.8 \\
VisionZip~\cite{yang2025visionzip} & 10\% & 53.5 & 58.0 & 42.5 & 49.3 & 53.4 & 51.3 & 91.2 \\
LLaVA-Scissor~\cite{sun2025llavascissor} & 10\% & - & 57.5 & - & - & 55.8 & - & - \\
DyToK~\cite{lilessfull} & 10\% & 55.3 & 58.4 & 42.4 & 52.3 & 54.7 & 52.6 & 93.5 \\
FastVID~\cite{shen2025fastvid}  & 10\% & 55.9 & 58.7 & 42.6 & 56.3 & \bf 57.3 & 54.2 & 96.2 \\
HoliTom~\cite{shao2025holitom}  & 10\%/5\% & 57.3 & \bf 61.2 & \underline{45.1} & 56.3 & \underline{56.8} & \underline{55.3} & \underline{98.3} \\
\bf Ours (w/o M)  & 10\% & \underline{57.4} & 60.5 & 44.7 & \bf 57.4 & 56.6 & \underline{55.3}  & \underline{98.3} \\
\bf Ours & 10\%/5\% & \bf 57.7 & \underline{60.6} & \bf 45.5 & \underline{56.4} & 56.6 & \bf 55.4 & \bf 98.4 \\

\midrule

FastV~\cite{chen2024image}  & 100\%/5\% & 51.2 & 53.9 & 35.8 & 47.9 & 49.7 & 47.7 & 84.7 \\
VisionZip~\cite{yang2025visionzip}  & 5\% & 45.2 & 51.9 & 37.4 & 46.4 & 48.2 & 45.8 & 81.4 \\
LLaVA-Scissor~\cite{sun2025llavascissor}  & 5\% & - & 56.6 & - & - & 53.3 & - & - \\
DyToK~\cite{lilessfull} & 5\% & 50.0 & 53.9 & 40.8 & 49.1 & 50.6 & 48.9 & 86.8 \\
FastVID~\cite{shen2025fastvid}  & 5\% & 53.0 & 57.1 & 42.1 & 51.1 & 54.2 & 51.5 & 91.5 \\
HoliTom~\cite{shao2025holitom}  & 5\%/2.5\% & \underline{55.6} & \bf 60.5 & 40.6 & 53.6 & 54.2 & 52.9 & 94.0 \\
\bf Ours (w/o M)  & 5\% & \bf 56.4 & 59.5 & \underline{42.5} & \underline{53.7} & \bf 55.0 & \underline{53.4} & \underline{94.9} \\
\bf Ours  & 5\%/2.5\% & \bf 56.4 & \underline{60.2} & \bf 42.8 & \bf 54.5 & \underline{54.7} & \bf 53.7 & \bf 95.4 \\

\midrule

FastV~\cite{chen2024image} & 100\%/2\% & 49.0 & 50.6 & 34.1 & 47.1 & 47.3 & 45.6 & 81.0 \\
VisionZip~\cite{yang2025visionzip}  & 2\% & 41.7 & 47.6 & 31.8 & 45.1 & 45.9 & 42.4 & 75.3\\
DyToK~\cite{lilessfull} & 2\% & 42.8 & 48.4 & 33.3 & 45.9 & 47.1 & 43.5 & 77.3 \\
FastVID~\cite{shen2025fastvid}  & 2\% & 48.0 & 52.3 & 37.6 & 47.3 & 49.2 & 46.9 & 83.3 \\
HoliTom~\cite{shao2025holitom}  & 2\%/1\%  & 52.6 & \underline{57.2} & 37.4 & 48.5 & 51.1 & 49.4 & 87.7 \\
\bf Ours (w/o M)  & 2\% & \bf 52.9 & \underline{57.2} & \underline{39.5} & \bf 51.0 & \underline{51.3} & \underline{ 50.4 }& \underline{89.5}\\
\bf Ours  & 2\%/1\% & \underline{52.8} & \bf 57.6 & \bf 40.3 & \underline{50.8} & \bf 51.8 & \bf 50.7 & \bf 90.1 \\

\midrule
FastV~\cite{chen2024image} & 100\%/1\% & 48.2 & 48.8 & 32.3 & 45.5 & 46.2 & 44.2 & 78.5 \\
VisionZip~\cite{yang2025visionzip}  & 1\% & 40.8 & 43.8 & 29.7 & 44.4 & 44.3 &  40.6 & 72.1 \\
DyToK~\cite{lilessfull} & 1\% & 40.7 & 44.3 & 30.7 & 45.6 & 45.0 & 41.3 & 73.3 \\
FastVID~\cite{shen2025fastvid}  & 1\% & 45.3 & 47.8 & 32.4 & 46.1 & 47.0 & 43.7 & 77.7 \\
HoliTom~\cite{shao2025holitom}  & 1\%/0.5\% & \underline{49.6} &  52.9 & \underline{33.9} & 48.1 & 49.0 & 46.7  & 82.9 \\
\bf Ours (w/o M)  & 1\% & \bf 50.5 & \underline{53.3} & \bf 34.4 & \bf 49.1 & \bf 49.8 & \bf 47.4 & \bf 84.2 \\
\bf Ours  & 1\%/0.5\% & \bf 50.5 & \bf 53.8 & \bf 34.4 & \underline{48.8} & \underline{49.2} & \underline{47.3} & \underline{84.1} \\

\bottomrule
\end{tabular}
\end{table*}

\section{Results on Qwen2.5-VL-7B}

To evaluate the generality of our proposed method across different model architectures, we conduct additional experiments using the structurally distinct foundation model Qwen2.5-VL-7B~\cite{bai2025qwen2}. This model incorporates a vision encoder based on window attention, which enables dynamic adjustment of the input video frame resolution. As shown in \cref{tab:qwen}, our approach maintains strong performance on this architecture, achieving 85.5\% of the original model’s performance with only 2\% of total tokens, while substantially surpassing existing methods. These results further confirm the robust generalization capability of our method across diverse vision-language architectures.

\section{Result on ActivityNet-QA}
\begin{table}[t]
\caption{Comparison on ActivityNet-QA. The two metrics (\textit{Accuracy} and \textit{Score}) are sub-scores under the ActivityNet-QA benchmark.}
\centering
\setlength{\tabcolsep}{6pt}
\renewcommand{\arraystretch}{1.2} 
\begin{tabular}{l|c|cc}
\toprule
\multirow{2}{*}{\textbf{Method}} &
\multirow{2}{*}{\makecell{\bf Retain \\ \bf Ratio}} &
\multicolumn{2}{c}{\textbf{ActivityNet-QA}} \\
\cline{3-4}
 &  & \textbf{Accuracy} $\uparrow$ & \textbf{Score} $\uparrow$ \\
\midrule
\rowcolor{gray!20}
LLaVA-OV-7B & 100\% & 54.5 & 3.55 \\
+~VisionZip~\cite{yang2025visionzip}   & 2\% & 41.7 & 3.13 \\
+~FastVid~\cite{shen2025fastvid}     & 2\% &  49.6  & 3.35      \\
+~HoliTom~\cite{shao2025holitom}     & 2\% & 49.9 & 3.31 \\
+~Ours        & 2\% & \bf 50.2 & \bf 3.37 \\
\bottomrule
\end{tabular}

\label{tab:activitynetqa}
\end{table}
We also evaluate the proposed method on the ActivityNet~\cite{yu2019activitynet}, an open-ended question-answering benchmark. This task requires free-form answer generation, unlike multiple-choice video QA. Using GPT-3.5-Turbo as an evaluator, our approach is shown to outperform comparable methods in both accuracy and generative quality,as described in~\cref{tab:activitynetqa}.

\begin{table*}[t]
\centering
\caption{Comparison across different frame settings at 2\% token retention.}
\label{tab:high_frame_results}
\setlength{\tabcolsep}{4pt}
\begin{tabular}{l|c|ccccc|c}
\toprule
\makecell[c]{\bf Method}&
\makecell[c]{\bf Frames} &
\makecell[c]{\bf MVBench} &
\makecell[c]{\bf EgoSchema} &
\makecell[c]{\bf MLVU} &
\makecell[c]{\bf LongVideo \\ \bf Bench} &
\makecell[c]{\bf VideoMME} &
\makecell[c]{\bf Avg. Score} \\
\midrule
\rowcolor{gray!20}
Vanilla~\cite{li2024llava} & 16  & 58.1 & 59.5 & 41.8 & 55.7 & 56.7 & 54.4 \\
HoliTom~\cite{shao2025holitom} & 64  & 54.8 & 59.5 & 38.6 & 51.0 & 54.4 & 51.7 \\
Ours & 64  & 54.1 & 59.1 & 41.3 & 52.5 & 54.5 & 52.3 \\
HoliTom~\cite{shao2025holitom} & 96  & 55.9 & 60.4 & 41.0 & 51.4 & 54.1 & 52.6 \\
Ours & 96  & 55.7 & 59.5 & 42.9 & 54.2 & 55.9 & 53.6 \\
HoliTom~\cite{shao2025holitom} & 128 & 55.9 & \bf 60.9 & 41.3 & 53.8 & 55.4 & 53.5 \\
Ours    & 128 & \bf 56.1 & 59.5 & \bf 47.9 & \bf 55.2 & \bf 58.0 & \bf 55.3 \\
\bottomrule
\end{tabular}
\end{table*}

\section{Higher Frame Sampling Rates}

Our approach demonstrates robust performance advantages over Holitom at high frame sampling rates across key long-video benchmarks, including MLVU~\cite{zhou2025mlvu}, LongVideoBench~\cite{wu2024longvideobench}, and VideoMME~\cite{fu2025videomme}, as detailed in~\cref{tab:high_frame_results}. This consistent outperformance underscores its superior effectiveness for long-duration video understanding at ultra-low token retention.

\section{More Ablations}
\noindent\textbf{Ablation on pruning layer K within LLM.}
Ablation study in~\cref{tab:k_ablation_results} reveals that the effectiveness of the Text-Aware pruning strategy depends on its integration depth within the LLM. In shallow layers, performance is limited by insufficient alignment between visual features and textual semantics. In deeper layers, performance declines due to interference from unfiltered irrelevant visual information. The strategy performs optimally in intermediate layers, where visual-semantic integration is well-established and redundancy remains controllable, thereby enhancing the refinement of key information and improving task accuracy.

\begin{figure}[t!]
    \centering
    \includegraphics[width=1\linewidth]{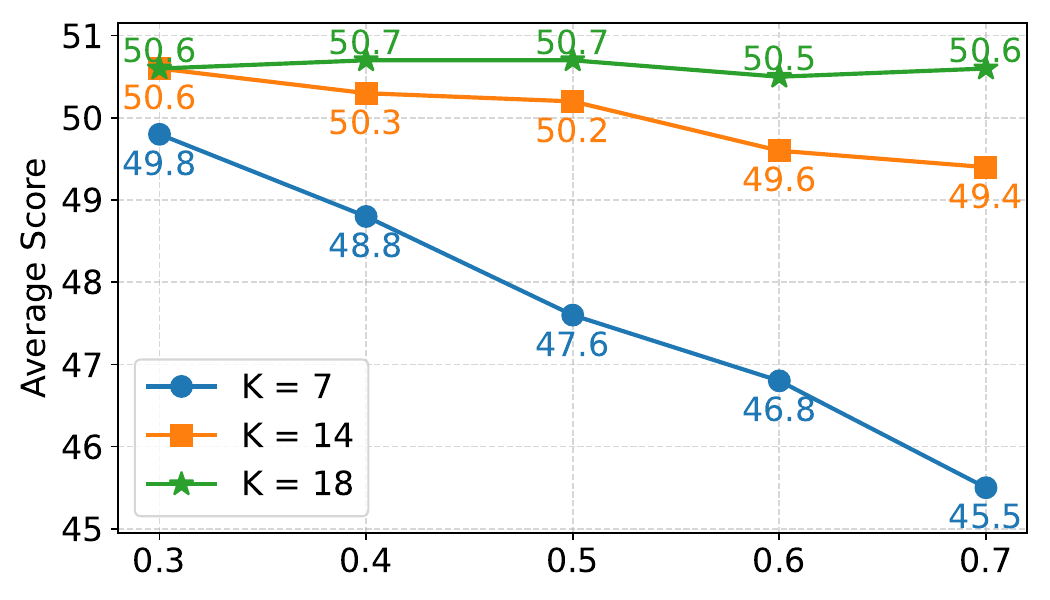}
    \caption{Performance of different pruning ratios across different layers}
    \label{fig:R}
\end{figure}

\noindent\textbf{Ablation on pruning ratio R within LLM.}
As illustrated in~\cref{fig:R}, different pruning ratios have distinct effects across network layers. While aggressive pruning in early layers substantially reduces FLOPs, it leads to sharp performance degradation. In contrast, deeper layers exhibit more aggregated visual features and increased token redundancy, making them more tolerant to higher pruning ratios. An effective balance between accuracy and computation therefore requires selecting appropriate pruning ratios according to layer depth.
\begin{table*}[t]
\centering
\caption{Performance comparison of different pruning layer K.}
\label{tab:k_ablation_results}
\begin{tabular}{c|c|ccccc|cc}
\toprule
\multirow{2}{*}{\bf Method}&
\multirow{2}{*}{\makecell{\bf K}} &
\multirow{2}{*}{\bf MVBench} &
\multirow{2}{*}{\bf EgoSchema} &
\multirow{2}{*}{\bf MLVU} &
\multirow{2}{*}{\makecell{\bf LongVideo \\ \bf Bench}} &
\multirow{2}{*}{\bf VideoMME} &
\multicolumn{2}{c}{\bf Avg.} \\
& & & & & & & \bf Score & \bf \% \\
\midrule
\rowcolor{gray!20}
Vanilla~\cite{li2024llava}& - & 58.3 & 60.4 & 47.7 & 56.4 & 58.6 & 56.3 & 100 \\ 

Ours & 2  & 48.7 & 52.7 & 36.7 & 47.8 & 49.4 & 47.1 & 83.6 \\
Ours & 7  & 50.3 & 54.0 & 36.6 & 47.8 & 49.5 & 47.6 & 84.6 \\
Ours & 14 & \bf 52.9 & 55.7 & 39.7 & \bf 51.1 & 51.4 & 50.2 & 89.1 \\
Ours & 18 & 52.8 & \bf 57.6 & \bf 40.3 &  {50.8} & \bf 51.8 & \bf 50.7 & \bf 90.1 \\
Ours & 21 &  {52.8} &  {57.1} &  {39.9} & 50.3 &  {51.7} &  {50.4} &  {89.5} \\
\bottomrule
\end{tabular}
\end{table*}
\section{Detailed Hyperparameter Analysis}
All experiments are conducted under a unified configuration described in Section~4.1. 
We provide a comprehensive analysis of key hyperparameters, including $\tau$, $\lambda$, and the clustering ratio, with results summarized in \cref{tab:tau_re,tab:lambda_re,tab:cluster_ratio}.
The selected values are chosen based on their overall performance across different settings, ensuring a stable and robust configuration. These results further validate the sensitivity and effectiveness of each component in our method.
\begin{table*}[h]
\centering
\caption{Comparison of $\tau$.}
\label{tab:tau_re}
\begin{tabular}{c|c|ccccc|cc}
\toprule
\multirow{2}{*}{\bf Method}&
\multirow{2}{*}{\makecell{\bf $\tau$}} &
\multirow{2}{*}{\bf MVBench} &
\multirow{2}{*}{\bf EgoSchema} &
\multirow{2}{*}{\bf MLVU} &
\multirow{2}{*}{\makecell{\bf LongVideo \\ \bf Bench}} &
\multirow{2}{*}{\bf VideoMME} &
\multicolumn{2}{c}{\bf Avg.} \\
& & & & & & & \bf Score & \bf \% \\
\midrule
\rowcolor{gray!20}
Vanilla~\cite{li2024llava} & - & 58.3 & 60.4 & 47.7 & 56.4 & 58.6 & 56.3 & 100 \\

Ours & 0.5
& 49.6 & 52.9 & 39.4 & 49.1 & 52.3
& 48.7 & 86.4 \\

Ours & 0.6
& 51.7 & 55.9 & \textbf{40.0} & \ {50.7} & \textbf{52.7}
& 50.3 & 89.4 \\

Ours & 0.7
& \ {52.9} & \textbf{57.2} & 39.5 & \textbf{51.0} & 51.3
& \textbf{50.4} & \textbf{89.5} \\

Ours & 0.8
& \textbf{53.1} & 56.4 & 38.2 & 49.4 & 51.3
& 49.7 & 88.2 \\

Ours & 0.9
& 52.1 & 52.4 & 37.1 & 49.7 & 50.7
& 48.4 & 86.0 \\
\bottomrule
\end{tabular}
\end{table*}
\begin{table*}[h]
\centering
\caption{Comparison of Cluster Ratio at 2\% Tokens.}
\label{tab:cluster_ratio}

\begin{tabular}{c|c|ccccc|cc}
\toprule
\multirow{2}{*}{\bf Method}&
\multirow{2}{*}{\makecell{\bf Ratio}} &
\multirow{2}{*}{\bf MVBench} &
\multirow{2}{*}{\bf EgoSchema} &
\multirow{2}{*}{\bf MLVU} &
\multirow{2}{*}{\makecell{\bf LongVideo \\ \bf Bench}} &
\multirow{2}{*}{\bf VideoMME} &
\multicolumn{2}{c}{\bf Avg.} \\
& & & & & & & \bf Score & \bf \% \\

\midrule
\rowcolor{gray!20}
Vanilla & - & 58.3 & 60.4 & 47.7 & 56.4 & 58.6 & 56.3 & 100 \\

Ours & 0.1
& 51.1 & 56.5 & 39.0 & 50.7 & 51.6
& 49.7 & 88.4 \\

Ours & 0.2
& 51.6 & 56.8 & 38.6 & 49.9 & 51.3
& 49.6 & 88.2 \\

Ours & 0.3
& \textbf{52.9} & \textbf{57.2} & \textbf{39.5} & 51.0 & 51.3
& \textbf{50.4} & \textbf{89.5} \\

Ours & 0.4
& 52.2 & 56.5 & 38.4 & 49.4 & \textbf{52.0}
& 49.7 & 88.3 \\

Ours & 0.5
& 52.4 & 56.5 & 38.7 & \textbf{51.2} & \textbf{52.0}
& 50.2 & 89.1 \\

Ours & 0.6
& 52.8 & 57.0 & 37.6 & 51.0 & 51.2
& 49.9 & 88.7 \\

Ours & 0.7
& 52.6 & 56.1 & 36.1 & 50.5 & 51.3
& 49.3 & 87.6 \\
\bottomrule
\end{tabular}
\end{table*}
\begin{table*}[h]
\centering
\caption{Comparison of $\lambda$.}
\label{tab:lambda_re}
\begin{tabular}{c|c|ccccc|cc}
\toprule
\multirow{2}{*}{\bf Method}&
\multirow{2}{*}{\makecell{\bf $\lambda$}} &
\multirow{2}{*}{\bf MVBench} &
\multirow{2}{*}{\bf EgoSchema} &
\multirow{2}{*}{\bf MLVU} &
\multirow{2}{*}{\makecell{\bf LongVideo \\ \bf Bench}} &
\multirow{2}{*}{\bf VideoMME} &
\multicolumn{2}{c}{\bf Avg.} \\
& & & & & & & \bf Score & \bf \% \\
\midrule
\rowcolor{gray!20}
Vanilla~\cite{li2024llava} & - & 58.3 & 60.4 & 47.7 & 56.4 & 58.6 & 56.3 & 100 \\

Ours & 0
& 52.6 & 57.5 & \textbf{40.4} & 50.6 & 51.5
& 50.5 & 89.7 \\

Ours & 0.25
& 52.6 & 57.5 & 40.1 & 50.7 & 51.5
& 50.5 & 89.7 \\

Ours & 0.5
& \textbf{52.8} & \textbf{57.6} & 40.3 & \textbf{50.8} & \textbf{51.8}
& \textbf{50.7} & \textbf{90.1} \\

Ours & 0.75
& 52.5 & 57.5 & 39.9 & 50.5 & 51.5
& 50.4 & 89.5 \\

Ours & 1
& 52.6 & \textbf{57.6} & 40.0 & 50.6 & 51.6
& 50.5 & 89.7 \\
\bottomrule
\end{tabular}
\end{table*}

\section{Visualizations}

\begin{figure*}[!htbp]
    \centering
    \includegraphics[width=\linewidth]{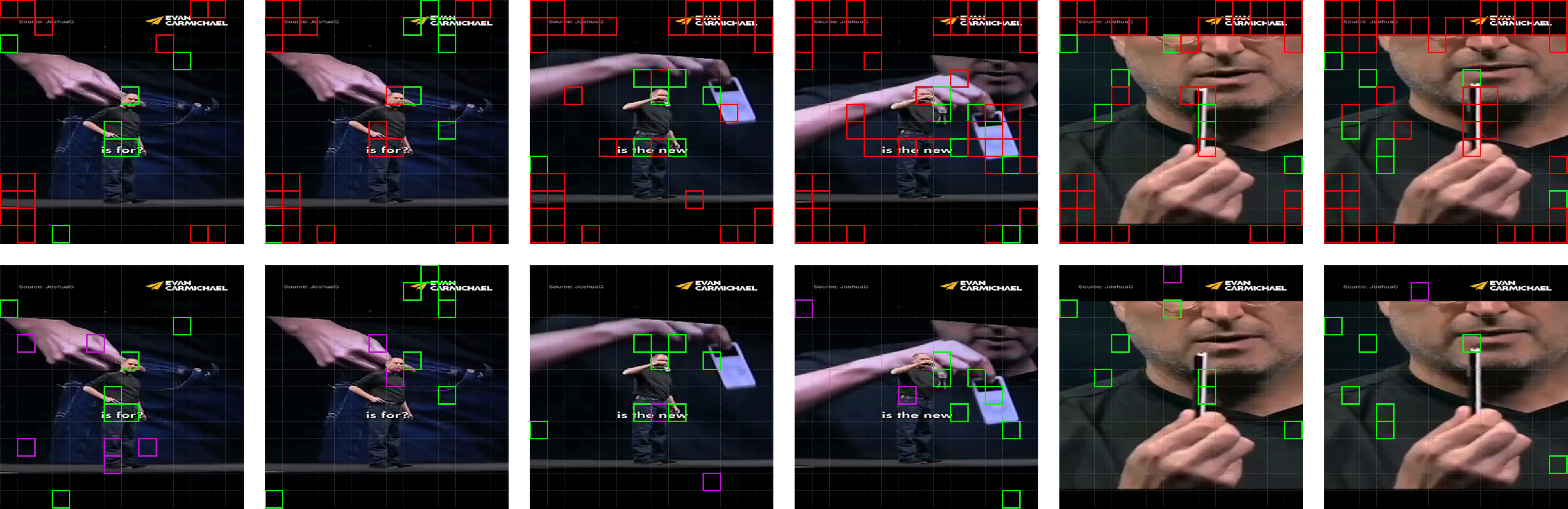}\\[1em]
    \includegraphics[width=\linewidth]{figs/visualization/pen.pdf}\\[1em]
    \includegraphics[width=\linewidth]{figs/visualization/obj.pdf}
\caption{
Token reductions outside the LLM. 
\textcolor{green!60!blue}{Green} boxes mark retained tokens, 
\textcolor{red}{Red} boxes mark tokens discarded by similarity filtering, and 
\textcolor{red!50!blue}{Purple} boxes mark tokens merged via clustering.
}
    \label{fig:outllm}
\end{figure*}

\noindent\textbf{Unified Spatiotemporal Compression.} 
The unified spatiotemporal compression process is visualized in~\cref{fig:outllm}. Red tokens indicate pruned redundant tokens transferred to the recycling pool through similarity-based pruning, whereas green tokens represent preserved key information. This process effectively eliminates spatial redundancy. Moreover, clustering tokens from the recycling pool yields purple tokens that serve as semantic supplements, maintaining full semantic integrity of the video content.

\begin{figure*}[t]
    \centering
    \includegraphics[width=0.99\linewidth]{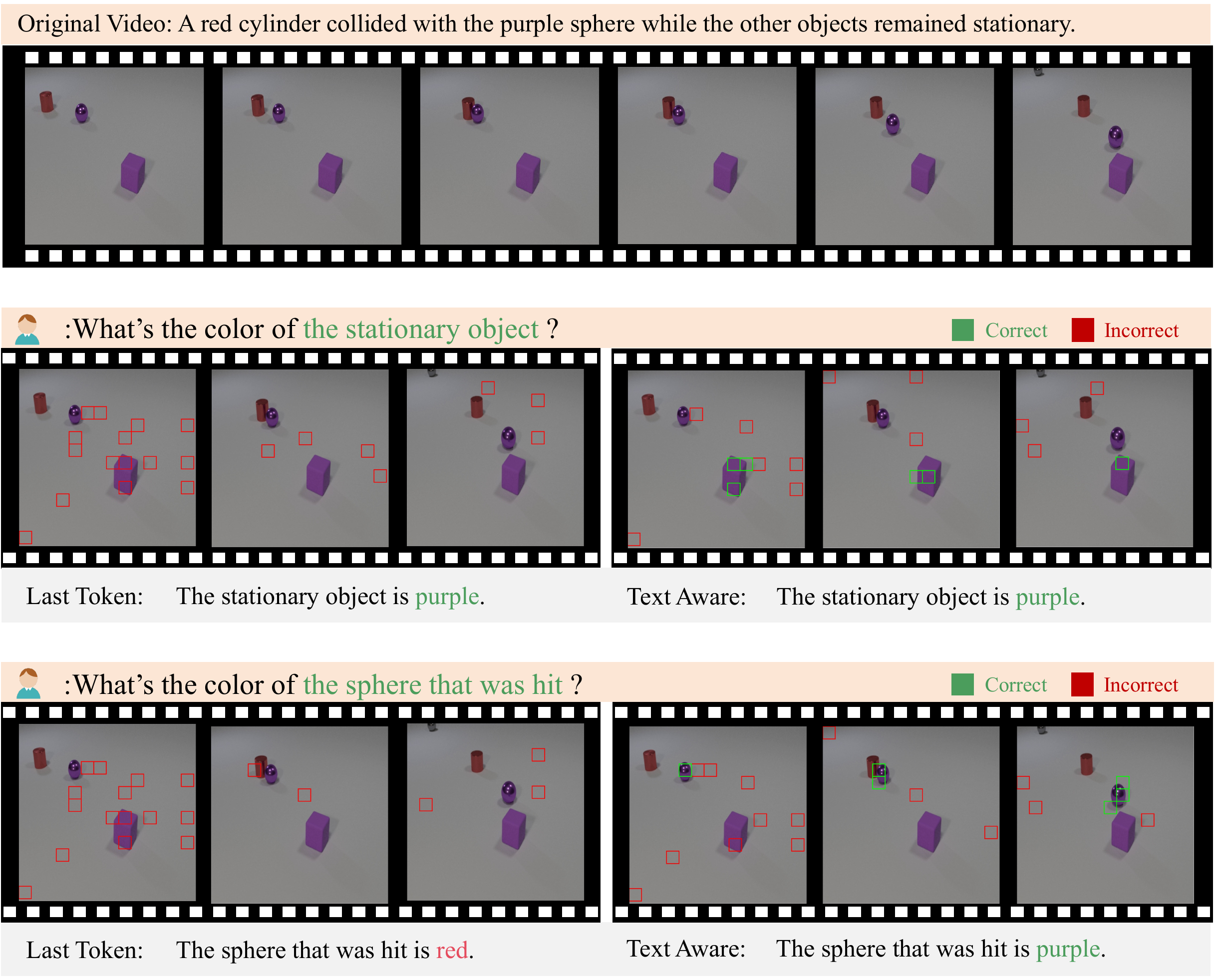}
    \caption{Tokens selected by our text-aware module within the LLM.}
    \label{fig:inllm}
\end{figure*}

\noindent\textbf{Text-aware Merging.}
\Cref{fig:inllm} illustrates the behavior of the text-aware mechanism in the LLM. For identical video inputs, our approach dynamically attends to question-relevant visual tokens while filtering out irrelevant information. This targeted filtering mechanism enhances the precision of the generated responses.

\section{Limitations and Future Work}
Despite its effectiveness in accelerating video LLM inference under ultra-low retention ratios, our approach has certain limitations. It currently supports only fixed offline videos and lacks real-time streaming token compression. Furthermore, uniform frame sampling can result in redundant frames for short videos while potentially missing critical segments in long videos. Future work will focus on jointly optimizing frame selection and token compression to enhance performance in extreme low-retention settings.

\begin{figure*}[t]
    \centering
    \includegraphics[width=\linewidth]{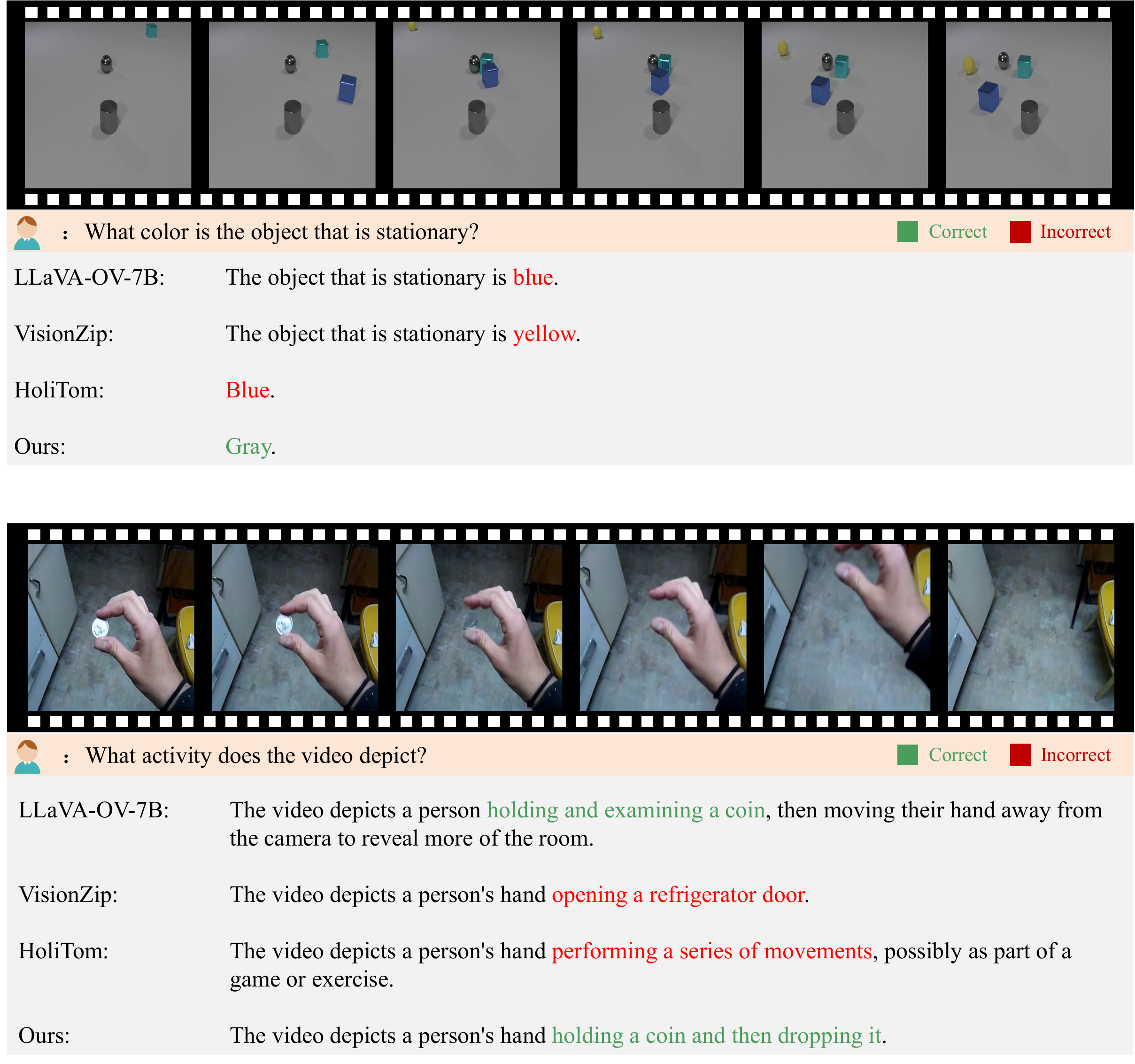}
    \caption{Qualitative generation comparison between our method and other approaches.}
    \label{fig:answer}
\end{figure*}

\end{document}